\documentclass[twoside]{article}

\usepackage[accepted]{aistats2023}
\usepackage[backend=biber,
style=apa,
maxcitenames=2,
maxbibnames=100, 
language=english,
doi=false,
isbn=false,
url=false,
uniquename=false
]{biblatex} 
\DeclareLanguageMapping{english}{english-apa} 
\usepackage{subcaption}
\usepackage{placeins}
\usepackage{mathtools}
\usepackage{amssymb}
\usepackage{amsmath}
\usepackage{amsfonts}   
\usepackage{fancyhdr}
\usepackage{times}
\usepackage{amsthm}
\usepackage{siunitx}
\usepackage{nicefrac}

\newtheorem{proposition}{Proposition}
\usepackage{todonotes}
\usepackage{algorithm}
\usepackage{algorithmic}

\newcommand{\M}{\mathcal{M}}
\newcommand{\N}{\mathcal{N}}
\newcommand{\T}{\mathfrak{T}}
\newcommand{\U}{\mathrm{Uniform}}
\newcommand{\0}{\boldsymbol{0}}
\newcommand{\X}{\mathbf{X}}

\newcommand{\diff}{\mathrm{d}}
\newcommand{\given}{\,|\,}

\newcommand{\summary}{\mathbb{T}}

\DeclareMathOperator{\diag}{diag}

\usepackage{float}
\usepackage{ctable}
\usepackage{stfloats}
\usepackage{accents}
\newcommand{\simulated}[1]{\protect\accentset{\text{s}}{#1}}
\newcommand{\observed}[1]{\protect\accentset{\text{o}}{#1}}

\usepackage{enumitem}

\usepackage{xcolor}
\definecolor{darkblue}{RGB}{0,0,128}
\definecolor{darkgreen}{RGB}{0, 128, 0}
\definecolor{darkred}{RGB}{128, 0, 0}
\definecolor{black}{RGB}{0, 0, 0}
\definecolor{errorcolor}{HTML}{481567}
\definecolor{viridisgreen}{HTML}{55C667}

\usepackage{forest}
\usepackage{adjustbox}

\usepackage[%
  colorlinks = true,
  citecolor  = darkblue,
  linkcolor  = darkblue,
  urlcolor   = darkblue,
  pdfauthor={Marvin Schmitt, Stefan T. Radev, Paul-Christian Bürkner},
  pdftitle={Meta-Uncertainty in Bayesian Model Comparison}
  ]{hyperref}

\begin{document}

%

%

\twocolumn[

\aistatstitle{Meta-Uncertainty in Bayesian Model Comparison}

\aistatsauthor{Marvin~Schmitt \And Stefan~T.~Radev \And Paul-Christian~Bürkner}

\aistatsaddress{%
Cluster of Excellence SimTech\\University of Stuttgart\\Germany \And 
Cluster of Excellence STRUCTURES\\Heidelberg University\\Germany \And
Cluster of Excellence SimTech\\University of Stuttgart\\Germany
}]

\begin{abstract}
Bayesian model comparison (BMC) offers a principled probabilistic approach to study and rank competing models.
In standard BMC, we construct a discrete probability distribution over the set of possible models, conditional on the observed data of interest.
These posterior model probabilities (PMPs) are measures of uncertainty, but---when derived from a finite number of observations---are also uncertain themselves.
In this paper, we conceptualize distinct levels of uncertainty which arise in BMC.
We explore a fully probabilistic framework for quantifying \textit{meta-uncertainty}, resulting in an applied method to enhance any BMC workflow.
Drawing on both Bayesian and frequentist techniques, we represent the uncertainty over the uncertain PMPs via \emph{meta-models}
which combine simulated and observed data into a predictive distribution for PMPs on new data.
We demonstrate the utility of the proposed method in the context of conjugate Bayesian regression, likelihood-based inference with Markov chain Monte Carlo, and simulation-based inference with neural networks.
\end{abstract}


\section{INTRODUCTION}
Scientific reasoning rarely comes with absolute certainty.
However, properly accounting for \emph{uncertainty} at each step of a research project is challenging, let alone quantifying and communicating all sources of uncertainty adequately.

Bayesian statistics offer an end-to-end probabilistic framework for assigning probability distributions to model parameters and propagating the associated uncertainty throughout the entire workflow \autocite{gelman_bayesian_2014}.

Yet, assigning probability distributions over parameters might not address every question of interest.
In fact, substantial research questions frequently revolve around comparing several competing models for an empirical phenomenon \autocite{etz_introduction_2018}.
Since any researcher can favor any model, assessing the quality of a given model is of utmost importance, and a vast number of metrics exist to evaluate a proposed model \autocite[e.g.,][]{konishi_information_2008, konishi_generalised_1996, akaike_new_1974}.
These metrics become crucial to applied statistical modeling when several models are proposed and need to be compared against each other \autocite{burkner_models_2022}.

As a recent example, various researchers have put forward models of disease outbreaks in order to provide forecasts and estimate the utility of interventions during the COVID-19 pandemic \autocite[e.g.,][]{wagner_2022_modelling, dehning_inferring_2020}.
Consequently, policy makers need to compare the relative merits of plausible candidate models and decide which models to consider as a basis for implementing effective policies.
Thus, model comparison (along with model selection) turns into a central topic with a potentially enormous societal impact.


\begin{figure*}[t]
    \centering
    \includegraphics[width=\linewidth]{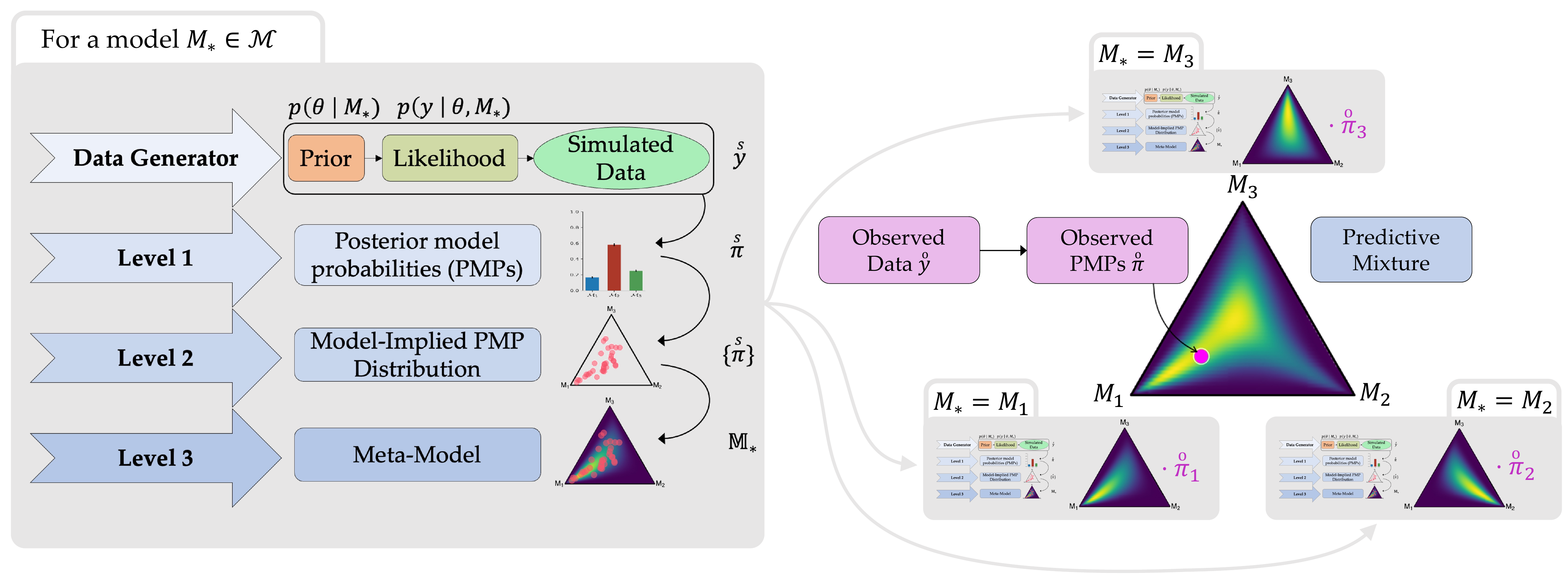}
    \caption{Meta-uncertainty acts as an overarching concept to capture the uncertainty in posterior model probabilities (PMPs), which are measures of uncertainty themselves.
    The result is a posterior predictive mixture distribution for PMPs on new observed data, for example from a replication study.
    This predictive mixture distribution is consistent, mitigates overconfidence, and gauges the reproducibility of observed model comparison results in future studies or experiments.
    }
    \label{fig:overview-meta-uncertainty}
\end{figure*}

The current work is located at the core of Bayesian model comparison research.
We study how distinct sources of uncertainty arise in Bayesian model comparison settings and introduce \emph{meta-uncertainty} as a novel concept to describe the model-implied variation of inference quantities (see \autoref{fig:overview-meta-uncertainty}).
Specifically, our main contributions are:
\begin{enumerate}[label=(\roman*), topsep=4pt,itemsep=8pt,partopsep=0pt, parsep=0pt]
    \item a fully probabilistic framework to describe qualitatively distinct \emph{levels of uncertainty} arising in any Bayesian model comparison setting; 
    \item a method to \emph{augment} observed data through simulated data, leading to a principled embedding of observed posterior model probabilities into the probabilistic context of all competing models;
    \item a formal argument that our method is \emph{consistent} and \emph{mitigates overconfidence} under certain conditions;
    \item an actionable interpretation for applied researchers in terms of the \emph{replicability} of observed results.
\end{enumerate}

\section{BACKGROUND}

\paragraph{Bayesian Statistics}
Bayesian methods constitute a principled approach to address questions of statistical inference from a self-consistent probabilistic perspective \autocite{gelman_bayesian_2014}.
Bayesian models formalize the assumed data generating process as
\begin{equation}\label{eq:bayesian-data generating-process}
	y\sim p(y\given\theta, M) \quad\text{with}\quad \theta\sim p(\theta\given M),
\end{equation}
with observed quantities (data)~$y\in\mathcal{Y}$, parameters~$\theta\in\Theta$, and model~$M\in\mathcal{M}$.
The posterior distribution 
\begin{equation}\label{eq:bayes-theorem-model}
    p(\theta\given y, M) = 
    \dfrac{p(y\given\theta, M)\,p(\theta\given M)}{p(y\given M)}
\end{equation}
is the main endpoint of Bayesian inference as it captures our information gain regarding the parameters~$\theta$ of model 
$M$, conditional on the data~$y$.

\paragraph{Uncertainty Quantification} In a Bayesian context, uncertainty is a paramount concept which often entertains distinct meanings and consequences \autocite{hullermeier_aleatoric_2021}.
The posterior distribution in \autoref{eq:bayes-theorem-model} implies that our knowledge about the parameter values~$\theta$ is not absolutely precise, but surrounded by \emph{epistemic} uncertainty.
Notably, epistemic uncertainty is reducible: Under certain regularity conditions, the posterior distribution~$p(\theta\given y, M)$ will shrink to a Dirac impulse, with no epistemic uncertainty left as the number of observations approaches infinity \autocite{van_der_vaart_asymptotic_2000}.

However, even if the exact ``true'' values of the parameters~$\theta$ were known and there were no remaining epistemic uncertainty, \autoref{eq:bayesian-data generating-process} would still entail uncertainty due to the likelihood function $p(y\given\theta, M)$.
This remaining uncertainty is irreducible and follows from the \emph{aleatoric} ``intrinsic randomness'' of the data generating process \autocite{hullermeier_aleatoric_2021}.

\paragraph{Bayesian Model Comparison} In light of observed data~$y$, Bayesian model comparison (BMC) establishes a framework to compare $J$ candidate models~$M_1,\ldots,M_J\in\mathcal{M}$ conditional on the observations~$y$. 
A key quantity in BMC is the marginal likelihood
\begin{equation}\label{eq:marginal-likelihood-definition}
	p(y\given M_j) = \int_{\Theta_j} p(y\given\theta, M_j)\,p(\theta\given M_j)\diff\theta,
\end{equation}
which quantifies the evidence for model $M_j$ by averaging the likelihood function over all possible values of the parameters~$\theta$.

Once we have obtained (or approximated) the marginal likelihoods of individual models, we can compute their \emph{posterior model probabilities} (PMPs) via Bayes' rule
\begin{equation}\label{eq:pmp-definition}
	\pi_j := p(M_j\given y) = \dfrac{p(y\given M_j)p(M_j)}{\sum\limits_{M'\in\M} p(y\given M')\,p(M')},
\end{equation}
where~$p(M)$ represents the prior model probability (e.g., uniform if models are deemed equally plausible before observing any data). 
To shorten notation, we collect the PMPs of all considered models into a vector~$\pi=(\pi_{1}, \ldots, \pi_{J})$.

By definition, PMPs \autocite[as well as Bayes Factors;][]{kass_bayes_1995} are sensitive to the specification of the prior \autocite{oelrich_when_2020, schad_workflow_2021}, and specifying sensible joint priors is extremely challenging in practical modeling scenarios \autocite{lindley_statistical_1957, van_dongen_prior_2006, zitzmann_prior_2021, dellaportas_joint_2012, aguilar_intuitive_2022}.
Consequently, our method for augmenting PMPs can help practitioners evaluate their prior choices by unveiling undesired patterns in the resulting PMPs for both (i) samples from each model's prior predictive distribution and (ii) actually observed real data.

\paragraph{Predictive Distribution} 
Differently from marginal likelihood approaches, we can also perform BMC based on posterior predictive performance.
This family of methods builds on the \emph{posterior predictive distribution}, which expresses the probability of new data~$\tilde{y}$ by accounting for all uncertainty implied by the posterior (cf.~\autoref{eq:bayes-theorem-model}):
	\begin{equation}\label{eq:posterior-predictive}
	    p(\tilde{y}\given y, M_j) = \int_{\Theta_j} p(\tilde{y}\given\theta, y, M_j)p(\theta\given y, M_j)\diff\theta.
	\end{equation}
Predictive methods, typically paired with cross-validation, are less sensitive to prior specification, but do not enjoy the consistency guarantees of Bayes Factors and PMPs either \autocite{vehtari_bayesian_2002, vehtari_pareto_2021, vehtari_survey_2012, burkner_approximate_2020}.\footnote{See also \textcite{fong2020marginal} for a connection between the marginal likelihood and exhaustive cross-validation.}

Importantly, our method combines aspects from both marginal likelihood and posterior predictive approaches to arrive at a consistent framework for a more nuanced uncertainty quantification in model comparison settings.

\section{METHOD}\label{sec:methods}

Drawing on both Bayesian and frequentist concepts, we propose a mathematically sound description of uncertainty levels and use the resulting framework to augment observed PMPs with meta-uncertainty (see \autoref{fig:overview-meta-uncertainty} and \autoref{tab:overview}).



\subsection{Model-Implied Data and PMP Distributions}

Each Bayesian model~$M_j\in\M$ implies a recipe for simulating data~$\simulated{y}$ by randomly sampling from its prior and likelihood (\autoref{eq:bayesian-data generating-process}).
Thus, every Bayesian model with a proper prior is \emph{generative}, and simulating data from each model~$M_j\in\M$ is straightforward.
The decorated symbol $\simulated{y}$ emphasizes that the data are simulated from the generative models and not actually observed.
This distinction is essential in the context of our framework, as it integrates simulated and actually observed data.
By applying \autoref{eq:marginal-likelihood-definition} and \autoref{eq:pmp-definition}, we can calculate a vector of PMPs 
\begin{equation}
	\simulated{\pi} = 
	\Big(
		\simulated{\pi}_1,
		\ldots,
		\simulated{\pi}_J
	\Big) =
	\Big( 
		p(M_1\given \simulated{y}),
		\ldots,
		p(M_J\given \simulated{y})
	\Big)
\end{equation}
for any fixed simulated data set~$\simulated{y}$.
The resulting PMP vector~$\simulated{\pi}$ forms the basis of all subsequently discussed uncertainty levels.

\paragraph{Level 1 Uncertainty}
Each model~$M_j\in\M$ implies a marginal likelihood of the data~$y$, which integrates out the parameters~$\theta$ from the joint distribution $p(\theta, y)$ and thus averages out the epistemic uncertainty encoded in the prior distribution (cf. \autoref{eq:marginal-likelihood-definition}).
The uncertainty in the PMP vector---i.e., deviations from a one-hot encoding (aka.\ dummy coding with $J$ instead of $J{-}1$ variables) where one PMP equals one and all others equal zero---is essential to our framework and we refer to it as \emph{level 1 uncertainty}.
This uncertainty is epistemic, because we could, in principle, reduce it with more data:
In the limit of an arbitrary number of observations, $N\to\infty$, the level 1 uncertainty vanishes,
since PMPs are consistent \autocite{barron_consistency_1999}.

\paragraph{Simulation Paradigm} By applying \autoref{eq:bayesian-data generating-process} repeatedly, we can generate a set of $K$ simulated data sets~$\{\simulated{y}^{(k)}\}_{k=1}^K$, which we call ``model-implied'' data.
This set of data sets corresponds to a sample from the prior predictive distribution.
Computing the PMPs for each simulated data set in $\{\simulated{y}^{(k)}\}_{k=1}^K$ leads to a set of ``model-implied'' PMPs~$\{\simulated{\pi}^{(k)}\}_{k=1}^K$.
\autoref{alg:model-implied-pmp-distribution} outlines the essential steps of these computations.
\autoref{fig:experiment-1-level-2-level-3} shows the distribution of model-implied PMPs under each candidate model for a toy example in \textbf{Experiment 1}.

\begin{algorithm}[t]
\caption{Generation of a model-implied PMP distribution (level 2 uncertainty).}
\label{alg:model-implied-pmp-distribution}
\begin{algorithmic}[1]
\REQUIRE{$\M, K, N$}\hfill
\FOR{$k=1,\ldots,K$}
	\STATE{$M_*^{(k)}\sim p(M)$} \hfill\COMMENT{sample true model}
	\STATE{$\theta^{(k)} \sim p(\theta\given M_*^{(k)})$}\hfill\COMMENT{sample parameter}
	\STATE{$\simulated{y}^{(k)}\sim p(y\given\theta^{(k)},M_*^{(k)})$}\hfill\COMMENT{sample $N$ observations}
	\STATE{$\simulated{\pi}^{(k)} \leftarrow \big( 
		p(M_1\given \simulated{y}^{(k)}),
		\ldots,
		p(M_J\given \simulated{y}^{(k)})
		\big)$}\label{alg:model-implied-pmp-distribution:pmp-calculation}\hfill\COMMENT{PMPs}
\ENDFOR
\RETURN $\big\{(\simulated{\pi}^{(k)}, M_*^{(k)})\big\}_{k=1}^K$
\end{algorithmic}
\end{algorithm}

\paragraph{Level 2 Uncertainty} The data generating process outlined in \autoref{alg:model-implied-pmp-distribution} features aleatoric uncertainty, which translates into variation in the PMPs~$\simulated{\pi}^{(k)}$ across simulations from the same model.
We call the variation inherent in the \emph{model-implied} PMP distribution \emph{level 2 uncertainty}.
The level 2 uncertainty yields a notion of the aleatoric uncertainty in the model-implied distribution of PMPs due to the inherent randomness of the data generating process:
The PMPs conditional on two finite data sets~$\simulated{y}^{(1)}, \simulated{y}^{(2)}$ may differ even if the two data sets were generated by the same parameter configuration $\theta$ of the same true model~$M_{*}$.
The model-implied PMP distributions may unveil complex patterns and interactions between the models on an intuitive scale, that is, in terms of PMPs.
In practical applications, this may guide practitioners' endeavors for both model specification and design optimization.

\paragraph{Relation to Sampling Distributions}
Studying the statistical behavior of a sample of inference targets--- PMPs in our case---under competing models is closely related to the concept of sampling distributions, which aim to capture sample variations of frequentist origin \autocite{kulesa_sampling_2015, schad_workflow_2021}.
Formulating the statement ``model~$M_j$ is the true model~$M_{*}$'' as a hypothesis and the PMPs~$\pi$ as an inference target in accordance with frequentist terminology, the simulated sample of PMPs can be interpreted as a random draw from the sampling distribution~$p(\pi \given M_j=M_{*})$.
Consequently, as the number of simulations grows arbitrarily large ($K\to\infty$), we reproduce the \emph{exact} sampling distribution of the model-implied PMPs.
In what follows, we will utilize the model-implied distribution of PMPs to contextualize observed PMPs.

\paragraph{Computational Efficiency} Obtaining the model-implied PMPs (i.e., line~\ref{alg:model-implied-pmp-distribution:pmp-calculation} in \autoref{alg:model-implied-pmp-distribution}) can be quite demanding. 
On the one hand, computing the marginal likelihood for non-conjugate models is hard, since it requires integration.
In these most typical cases, we can resort to approximations based on Markov chain Monte Carlo (MCMC), such as bridge sampling \autocite{meng_warp_2002, gronau_bridgesampling_2020}.
On the other hand, the marginal likelihood becomes doubly intractable if the likelihood function itself is not available in a closed form \autocite[e.g., as in simulation-based inference,][]{cranmer_frontier_2020}.
In this case, we can fall back on approximate Bayesian computation \autocite{csillery_2010_approximate} or more scalable approximators relying on amortized neural estimation \autocite{radev_amortized_2021, radev_jana_2023}.
Indeed, in our experiments, we demonstrate the integration of meta-uncertainty across all three settings (i.e., conjugate, likelihood-based, and simulation-based).

\subsection{Meta-Models of Model-Implied PMP Distributions}
\label{sec:pmp-meta-models}
In this section, we will analyze the model-implied distribution of PMPs using a Bayesian approach.
First, we group the model-implied PMPs obtained via \autoref{alg:model-implied-pmp-distribution} by the respective true model indices~$M_*\in\M$, resulting in $J$ groups.
The indices~$j=1,\ldots,J$ express which model generated the data in the respective group.

For each group, we fit a Bayesian model with parameters~$\tau_j$ on the vectors of simulated PMPs~$\simulated{\pi}$ for data from the respective true model $M_j=M_*$, namely the set~$\{\simulated{\pi}^{(k)}\}_{M_*^{(k)}=M_j}$.
We refer to these models as \emph{meta-models} and denote them as $\mathbb{M}_j$.
To shorten notation, we abbreviate the set by writing  $\{\simulated{\pi}\}_j$ when the context is clear and no strictly formal notation is necessary.
The likelihood~$p_j(\pi \given \tau_j)$ of each meta-model $\mathbb{M}_j$ can be any density with support on the unit simplex, such as a Dirichlet density with $\tau=\alpha$, or a logistic Normal density with $\tau=(\mu, \Sigma)$.

Each meta-model~$\mathbb{M}_j$ yields a posterior distribution $p_j(\tau_j\given\{\simulated{\pi}\}_j)$ over its parameters~$\tau_j\in\mathfrak{T}_j$.
This posterior is informed by the sampling distribution of PMPs estimated from the corresponding model-implied data.
In practice, we can readily fit the meta-models with the R package \texttt{brms} \autocite{burkner_brms_2017,burkner_advanced_2018} as an interface to \texttt{Stan} \autocite{stan_development_team_stan_2022} to obtain $D$ draws $\{\tau_j^{(d)}\}_{d=1}^D$
from the posterior distribution~$p_j(\tau_j\given\{\simulated{\pi}\}_j)$ based on Hamiltonian Monte Carlo \autocite[HMC;][]{neal_mcmc_2011, betancourt_conceptual_2017}.
To shorten notation, we will refer to the $D$ posterior draws as~$\{\tau_j^{(d)}\}$.
Finally, each meta-model implies a posterior predictive distribution,
\begin{equation}\label{eq:predictive-mixture-posterior-predictive}
p_j(
\tilde{\pi}\given\{\simulated{\pi}\}_j
) = 
\int_{\T_j}
    p_j(
    \tilde{\pi}\given 
    \tau_j
    )\,
    p(
    \tau_j\given \{\simulated{\pi}\}_j
    )
    \diff\tau_j,
\end{equation}
which captures the distribution of PMPs for a new observed data set from the perspective of the meta-model $\mathbb{M}_j$.

\paragraph{Level 3 Uncertainty}

The $J$ posterior distributions of the meta-models have epistemic uncertainty owing to the finite number of simulated data sets $K$.
We refer to this uncertainty as \emph{level 3 uncertainty}, which can be reduced with more simulations and vanishes for $K\to\infty$.
What is more, the epistemic level 3 uncertainty is contingent on the aleatoric uncertainty from the meta-models' likelihood functions.
Thus, a substantial level 3 uncertainty might indicate that more simulations are required to improve the meta-model and learn about the patterns on level 2.

\begin{figure*}%
    \centering%
    \begin{subfigure}[t]{0.24\linewidth}%
        \includegraphics[width=\linewidth]{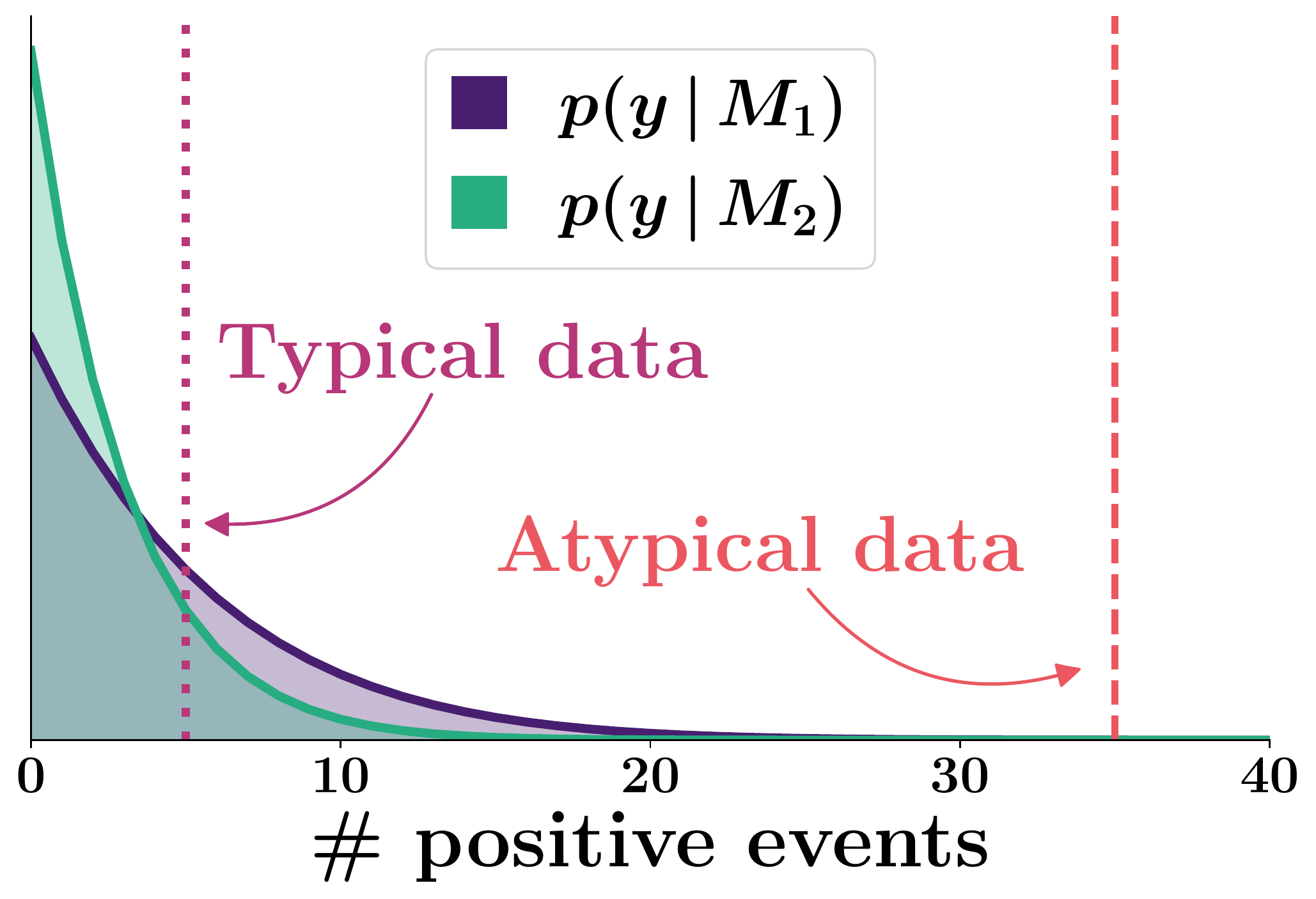}%
        \caption{Marginal likelihoods.}%
        \label{fig:minimal-example-pmp-ml:ml}%
    \end{subfigure}%
    \begin{subfigure}[t]{0.24\linewidth}%
        \includegraphics[width=\linewidth]{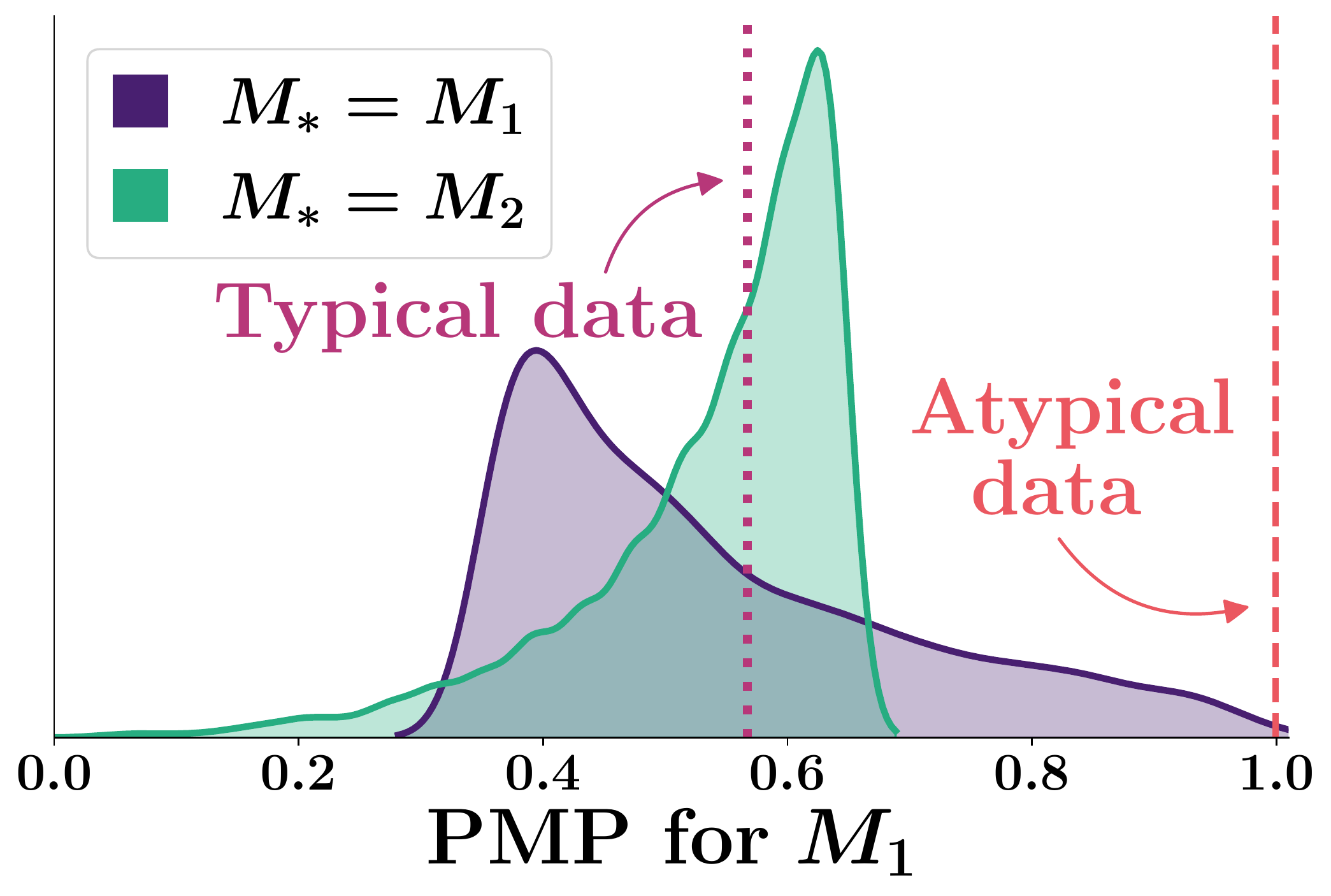}%
        \caption{Level 2 uncertainty.}%
        \label{fig:minimal-example-pmp-ml:pmp}%
    \end{subfigure}%
    \begin{subfigure}[t]{0.24\linewidth}%
        \includegraphics[width=\linewidth]{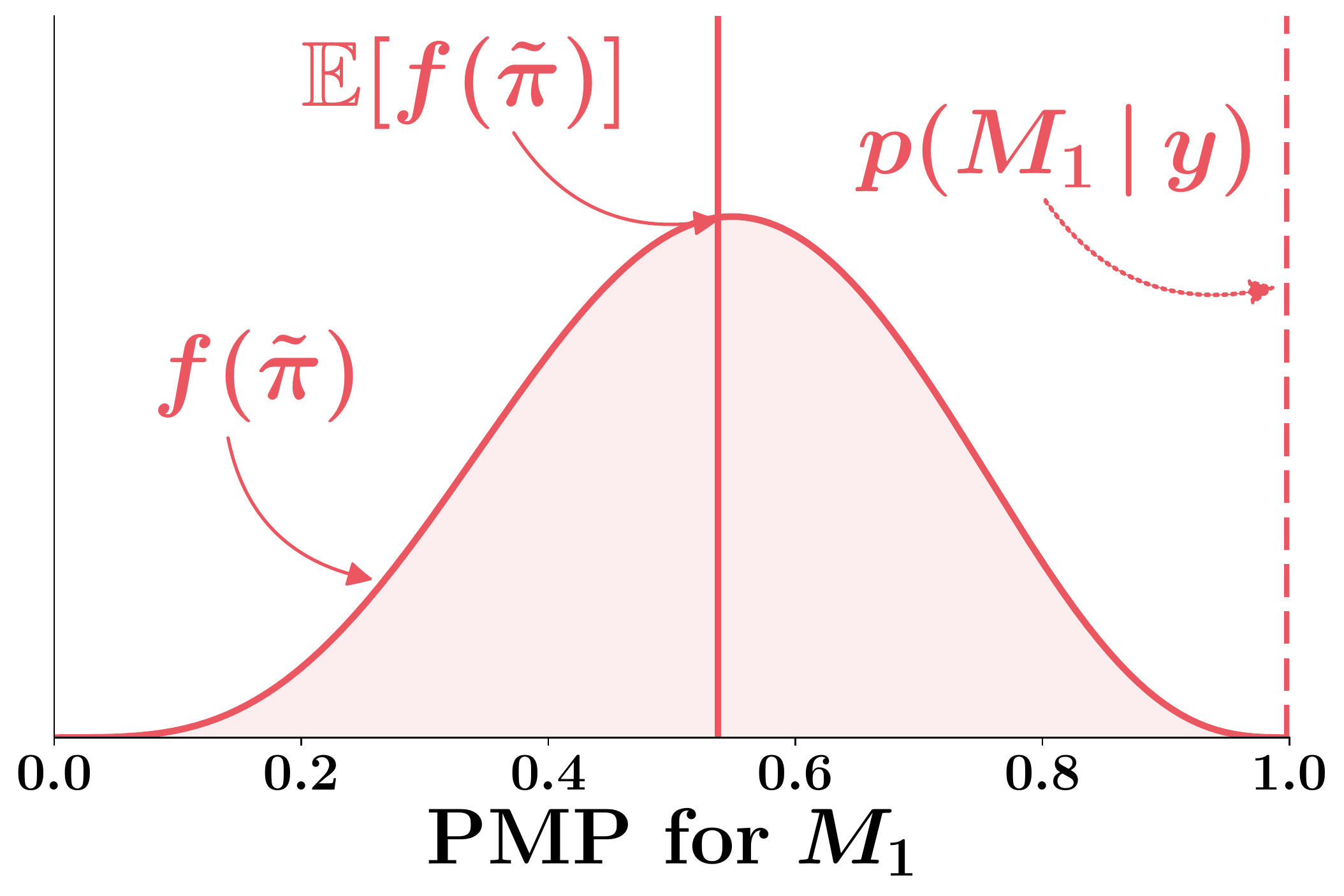}%
        \caption{Pred. mixture for atypical $y$.}%
        \label{fig:minimal-example-pmp-ml:predictive-mixture-atypical}%
    \end{subfigure}%
    \begin{subfigure}[t]{0.24\linewidth}%
        \includegraphics[width=\linewidth]{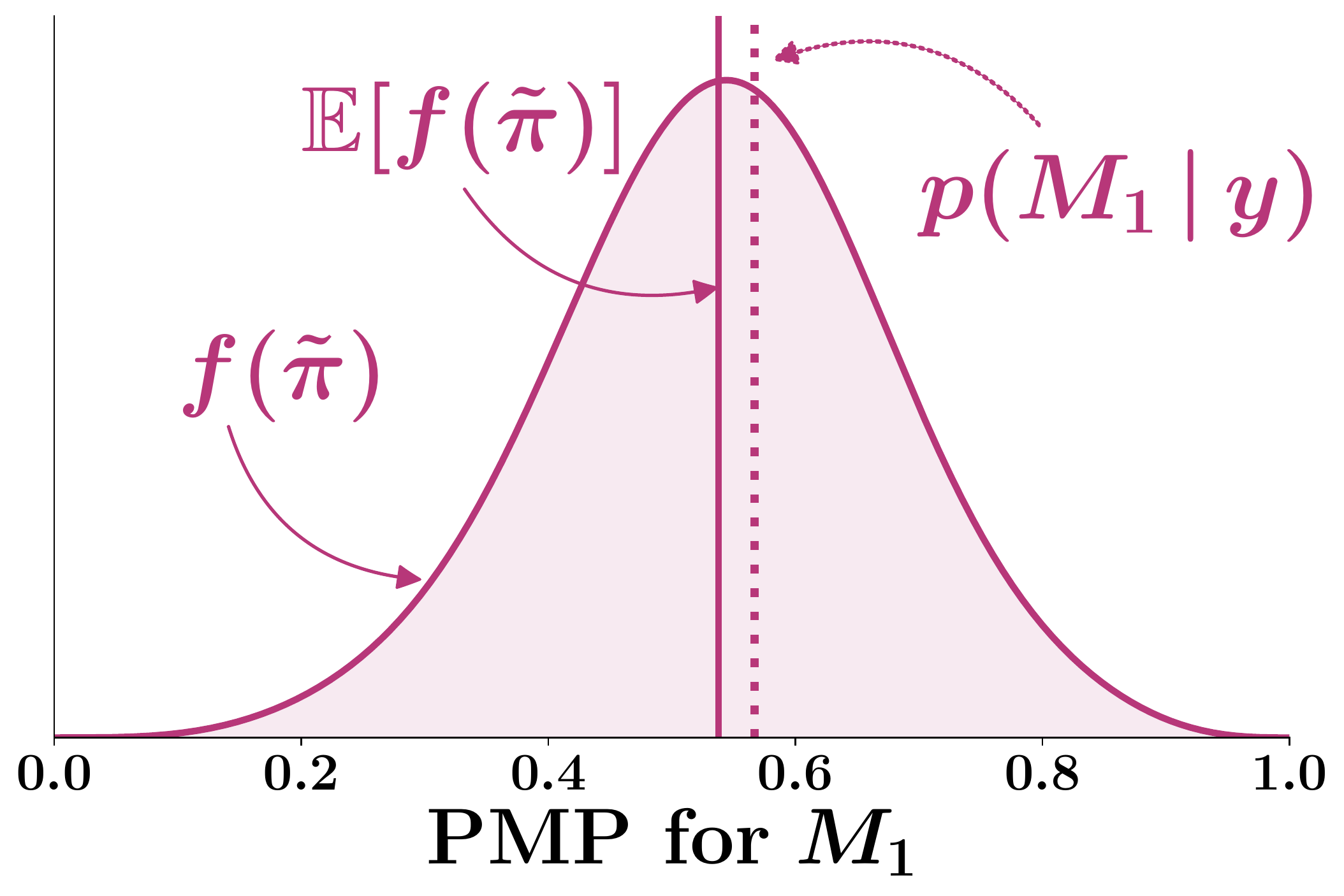}%
        \caption{Pred. mixture for typical $y$.}%
        \label{fig:minimal-example-pmp-ml:predictive-mixture-typical}%
    \end{subfigure}%
    \caption{Illustration of the predictive mixture method for the minimal motivating example comparing two conjugate Beta-Binomial models. Level 2 uncertainty in
    (\subref{fig:minimal-example-pmp-ml:pmp}) is visualized by a kernel density estimate (KDE) on $K=10\,000$ draws from the model-implied PMP distribution. The KDE is not part of our method but only used for visualization.}%
    \label{fig:minimal-example-pmp-ml}%
\end{figure*}%

\subsection{Predictive Mixture Distributions}\label{sec:predictive-mixture-construction}

We now describe how to use the information from model simulations and model-implied PMPs to inform the analysis of real data. 
For an actually observed data set~$\observed{y}$ and the set of candidate models~$\M=\{M_1,\ldots,M_J\}$, we first compute the observed PMPs~$\observed{\pi}=
(\observed{\pi}_1, \ldots,\observed{\pi}_J)$ according to \autoref{eq:pmp-definition}.
Then, we can embed the observed PMPs into the context of the model-implied PMPs distributions.
We achieve this by constructing a mixture distribution, whose components are the posterior predictive distributions of the meta-models from \autoref{eq:predictive-mixture-posterior-predictive} and whose mixture weights are the observed PMPs~$\observed{\pi}$:
\begin{equation}\label{eq:predictive-mixture-full}
\begin{aligned}
  f(\tilde{\pi}) = 
  \sum\limits_{j=1}^J
    \observed{\pi}_j \,
    p_j(\tilde{\pi}\given\{\simulated{\pi}\}_j).      
\end{aligned}
\end{equation}
Accordingly, $f(\tilde{\pi})$ represents a data-informed mixture distribution of simulation-based meta-models.
In other words, \autoref{eq:predictive-mixture-full} expresses a distribution over PMPs for a new (e.g., future or held-out) data set, based on the predictive distributions of the meta-models.
In real observational studies or experiments, the predictive mixture distribution can thus provide a \emph{notion of the reproducibility or replicability of a given model comparison result with regard to a subsequent study.}

In most analyses, the predictive distributions from \autoref{eq:predictive-mixture-posterior-predictive} and \autoref{eq:predictive-mixture-full} will be available to modelers as a set of (potentially thousands of) posterior draws.
Each posterior draw thus leads to a separate \emph{pushforward likelihood} and a separate density plot.
To ease visualization and analysis, we propose to
compute embeddings (summary statistics) of the posterior draws~$\{\tau_j^{(d)}\}$ with a set of $C$ suitable embeddings~$\{\summary_1\dots\summary_C\}$, $\mathbb{T}_c : \T^D \rightarrow \T$, leading to the computationally more feasible construction
\begin{equation}\label{eq:predictive-mixture}
    f_{\mathbb{T}_c}(\tilde{\pi}) = \sum\limits_{j=1}^J
    \observed{\pi}_j \,
    p_j\Big(
    \tilde{\pi}\,\Big|\,
    \summary_c
    \big(
    \{\tau_j^{(d)}\}
    \big)
    \Big).
\end{equation}
While the expected value would seem like a natural choice for $\summary$ with $C{=}1$, it might not behave as expected for multivariate data.\footnote{Pun intended. The expected value might not even lie in the ``typical set'' of a distribution \autocite{betancourt_conceptual_2017}.}
Instead, we can choose functions~$\summary_c(\{\tau_j^{(d)}\})$ which extract $C$ clusters from the posterior draws and yield the center within each cluster \autocite[cf.][for a similar clustering approach in projective inference]{piironen_projective_2020}.
Thus, for a set of $D$ posterior draws~$\{\tau_j^{(d)}\}$, the clustering approach reduces the number of resulting plots by a factor of ${D}/{C}$.
While our experiments in the main paper feature a mean embedding ($\summary=\mathbb{E}$) due to space constraints, the \textbf{Supplementary Material} contains results from the clustering technique for all experiments.

\subsection{Implications and Guarantees}

\begin{proposition}[Consistency]\label{prop:consistency}
    For $K>0$ and $M_*\in\M$ we have $f(\tilde{\pi}) \xrightarrow{\mathrm{a.s.}} \mathbb{I}_{M_*}$ as $N\to\infty$.
\end{proposition}
For an arbitrarily large number of observations $N\to\infty$, the PMPs (simulated and observed) converge almost surely to one-hot encoded vectors, $\pi \xrightarrow{\mathrm{a.s.}} \mathbb{I}_{M_*}$ with $p(M\given y)=1$ for $M=M_*$ and 0 for all other models (i.e.,  level 1 uncertainty approaches zero).
Consequently, level 2 uncertainty vanishes since all model-implied PMPs under each model are equal.
Accordingly, the meta-models will collapse to Dirac distributions, which in turn imply no level 3 uncertainty.
Thus, both the observed PMP and the mixture components in \autoref{eq:predictive-mixture-full} approach $\mathbb{I}_{M_*}$.
Finally, the predictive mixture distribution for a data set~$\observed{y}$ from model $M_*$ equals $\mathbb{I}_{M_*}$, which is consistent by definition.
We prove \autoref{prop:consistency} in the \textbf{Supplementary Material} and illustrate it in \textbf{Experiment 1} (see \autoref{fig:experiment-1-level-2-level-3}, $N=100$, bottom row).

\begin{proposition}[Mitigating Overconfidence]\label{prop:overconfidence} For $f(\tilde{\pi})\neq\mathbb{I}_{M_j}$, the joint distribution $p(\tilde{\pi}, M_j) = \observed{\pi}_j \,
    p_j(\tilde{\pi}\given\{\simulated{\pi}\}_j)$ has non-vanishing variance even if $\observed{\pi}$ has no level 1 uncertainty.
\end{proposition}


In the case of spuriously high PMPs in favor of a model $M_j$ (i.e., $\observed{\pi}_j \approx 1$), the predictive mixture distribution selects the sampling distribution of the respective model $M_j$. 
Given that the corresponding mixture component $p_j(\tilde{\pi}\given\{\simulated{\pi}\}_j)$ has a non-vanishing variance, we mitigate overconfidence:
Even if $\observed{\pi}=\mathbb{I}_{M_j}$, the predictive mixture still contains all variance from level 2.
We prove \autoref{prop:overconfidence} in the \textbf{Supplementary Material} and illustrate it in \textbf{Experiment 1} (cf. \autoref{fig:experiment-1-predictive-mixture:c}).

\begin{table*}[!b]
\setlength{\tabcolsep}{0.5em}
    \caption{Overview of the meta-uncertainty framework for $J$ models $M_1, \ldots, M_J$. We simulate $K$ data sets $\{\simulated{y}^{(k)}\}$ with $N$ observations from each of the models' prior predictive distributions.}
    \label{tab:overview}
    \centering
    \begin{tabular}{l|l|l}
    \textbf{Level} & \textbf{Core concepts} & \textbf{Core implications} \\
    \specialrule{.1em}{0em}{0em}
         Level 1 & Posterior model probabilities $\simulated{\pi} = p(\simulated{y} \given M_*)$ & Consistency of PMPs: $N\to\infty\xrightarrow{\mathrm{a.s.}}\pi\to\mathbb{I}_{M_*}$ (Prop.~\ref{prop:consistency})\\[2pt]
         \specialrule{.01em}{0em}{0em}
         Level 2 & Model-implied PMP distributions $\{\simulated{\pi}^{(k)}\}$ & Approach PMP sampling distribution for $K\to\infty$\\[2pt]
         \specialrule{.01em}{0em}{0em}
         Level 3 & Meta models on PMP distributions $p_j(\tau_j\given\{\simulated{\pi}\}_j)$ & Epistemic uncertainty of meta models fades for $K\to\infty$\\[2pt]
          & Predictive mixture $ f(\tilde{\pi}) = \sum\observed{\pi}_j \,p_j(\tilde{\pi}\given\{\simulated{\pi}\}_j)$ & Retain level 2 uncertainty $\to$ ease overconfidence (Prop.~\ref{prop:overconfidence})
    \end{tabular}
\end{table*}

\subsection{Minimal Example}

In order to convey some intuition about the applications and benefits of our framework, we consider a minimal example with two competing toy models $M_1$ and $M_2$.
The models follow the Beta-Binomial conjugacy
\begin{equation}
\begin{aligned}
    \theta &\sim \text{Beta}(\alpha, \beta),\\
    y_n &\sim \text{Bernoulli}(\theta),
\end{aligned}
\end{equation}
with two different prior configurations: $M_1{:}\ \alpha{=}1, \beta{=}10$; $M_2{:}\ \alpha{=}1, \beta{=}20$.
These priors assume that positive events (i.e., $y_n = 1$) are rare and differ solely with regard to how much mass they assign to this assumption.
The marginal likelihoods and associated PMPs are analytic \autocite{salvatier_probabilistic_2016}.
Clearly, the resulting PMPs will contain level 1 uncertainty for a finite number $N$ of observations.

To obtain a notion of level 2 uncertainty, we use \autoref{alg:model-implied-pmp-distribution} to generate $K=10\,000$ draws from the two model-implied PMP distributions ($N=50$ for all simulated data sets) and display these draws in \autoref{fig:minimal-example-pmp-ml:pmp}.
We then fit two Beta meta-models to these PMP draws, which we will use for constructing the predictive mixture distribution according to \autoref{eq:predictive-mixture-full}. 
To simplify exposition, we will focus on interpreting the mean of the predictive distribution,
\begin{equation}
    \mathbb{E}[f(\tilde{\pi})] = \observed{\pi}_1\, \mathbb{E}[p_1(\tilde{\pi} \given \{\simulated{\pi}\}_1)] + \observed{\pi}_2\, \mathbb{E}[p_2(\tilde{\pi} \given \{\simulated{\pi}\}_2)],
\end{equation}
with $\observed{\pi}_2 = (1 - \observed{\pi}_1)$ since we only compare two models.
As the epistemic uncertainty over meta-model parameters (level 3) is negligible due to the large number of simulations, we can further simplify the above expectation by representing it as a linear transformation of the observed PMPs
\begin{equation}\label{eq:linear_t_probs}
\mathbb{E}[f(\tilde{\pi})] \approx
    \begin{pmatrix}
    \bar{\pi}_1 & 1 - \bar{\pi}_2\\
    1 - \bar{\pi}_1 & \bar{\pi}_2
    \end{pmatrix}
    \begin{pmatrix}
    \observed{\pi}_1\\
    \observed{\pi}_2
    \end{pmatrix},
\end{equation}
where $\bar{\pi}_1$ and $\bar{\pi}_2$ denote the corresponding means of the meta-models.
These means correspond to the \emph{average magnitude of model-implied PMPs which we expect under the assumption that each of the two models is true}.

Consider now what happens if, for instance, the data generating process undergoes a shift and positive events become much more likely (see \autoref{fig:minimal-example-pmp-ml:ml}). 
In this case, both models will be severely misspecified, but $M_1$ will completely ``dominate'' $M_2$, since it has a slightly fatter right tail.
As observed by \textcite{yang_bayesian_2018}, researchers would ideally want an indication that the models are misspecified, but instead, PMPs will converge to $1$ for the ``least misspecified'' model \autocite{barron_consistency_1999} and model selection based on PMPs will be markedly overconfident.

Indeed, assuming an atypical observed data set $\observed{y}$ with $35$ positive and $15$ negative events (cf. \autoref{fig:minimal-example-pmp-ml:ml}), the PMP for model $M_1$ is $\observed{\pi}_1 \approx 0.999$.
However, \autoref{eq:linear_t_probs} suggests a more modest preference for $M_1$ and reduces the observed PMP to the average model-implied PMP~$\bar{\pi}_1$, amounting to no more than $\bar{\pi}_1 \approx 0.55$ (cf. \autoref{fig:minimal-example-pmp-ml:predictive-mixture-atypical}).
In contrast, the PMPs obtained on a typical data set are close to the mean of the corresponding predictive distribution (cf. \autoref{fig:minimal-example-pmp-ml:predictive-mixture-typical}).
Thus, in this example, the predictive mixture acts as a ``simulation-based correction'' for overconfident model selection in the presence of atypical data.
In other settings, it can be used to predict the expected PMPs in an upcoming experiment or a replication study (see \textbf{Experiment 1}).

\section{RELATED WORK}

\paragraph{Overconfidence and Dance} Researchers have observed that PMPs tend to concentrate on one candidate model, even if other evaluation metrics (such as predictive performance or information criteria) do not indicate a strong preference for that model \autocite{oelrich_when_2020} or the candidate models are misspecified \autocite{yang_bayesian_2018}.
This phenomenon is called \emph{overconfidence} and amounts to spuriously selecting one of the compared models as the only probable model in the set $\mathcal{M}$ \autocite{oelrich_when_2020, yang_bayesian_2018}.
What is more, a moderate change in the data often suffices for the PMPs to shift and concentrate on another model.
This behavior---large jumps of PMPs when the data is slightly altered---motivates the term ``dance of Bayes factors'' \autocite{oelrich_when_2020}.
Accordingly, our predictive mixture method could reduce overconfidence through its ``simulation-based correction'' of observed PMPs and its stronger emphasis on uncertainty.

\paragraph{Higher-Order Uncertainty}

\textcite{cheeseman_1985_defense} argues that single probability assignments are (often) sufficient for reasoning and deciding under uncertainty. 
Nevertheless, various generalizations of probability have been suggested over the years, such as the theory of random sets \autocite{nguyen_2008_random, matheron_1975_random} or (generalized) evidence theory \autocite{deng_generalized_2015, shafer_mathematical_1976}.
Roughly speaking, these ideas conceptualize higher-order uncertainty as \emph{sets of distributions} or \emph{distributions over distributions} \autocite[see][]{hullermeier_aleatoric_2021}.
Our method is inspired by these ideas and follows the premise that a single probability assignment is not sufficient for all practical purposes, as witnessed, for example, by the dangers of overconfident model choice \autocite{yang_bayesian_2018, oelrich_when_2020}. 

\paragraph{Closed vs.\ Open World} Evidence theory \autocite{shafer_mathematical_1976, deng_generalized_2015} differentiates between a \emph{closed world} and an \emph{open world} assumption.
This terminology is commonly acknowledged in statistical inference \autocite{bernardo_bayesian_2009, yao_using_2018}, where it describes fundamentally different settings for model comparison.
In a closed world setting, the ``true'' data generating model~$M_{*}$ is assumed to be one of the considered models, $M_{*}\in\mathcal{M}$.
Differently, in the open world setting, the ``true'' model may not be a member of $\mathcal{M}$ or may not even exist.

In the context of deep learning, the relaxation of the closed world assumption is known as ``out-of-distribution detection''
\autocite{liang_enhancing_2020, hendrycks_baseline_2018, devries_learning_2018}.
In the Bayesian literature, the problem is framed as ``model misspecification'' and has been studied both in the context of standard and generalized Bayesian inference \autocite{grunwald_inconsistency_2017, thomas_diagnosing_2019, masegosa_2020_learning, knoblauch_2019_generalized, giummole_2019_objective} as well as in the context of simulation-based inference \autocite{alquier_concentration_2019, zhang_convergence_2020, frazier_model_2020, frazier_robust_2021, hermans_2021_averting, schmitt_detecting_2022}.

Our method can be viewed as a contribution to this growing literature, as it explores a way to integrate information from a closed world (i.e., simulations from a finite set of candidate models) and an open world (i.e., the real data obtained from a potentially unknown source).

\begin{figure}[t]
    \centering
    \includegraphics[width=\linewidth]{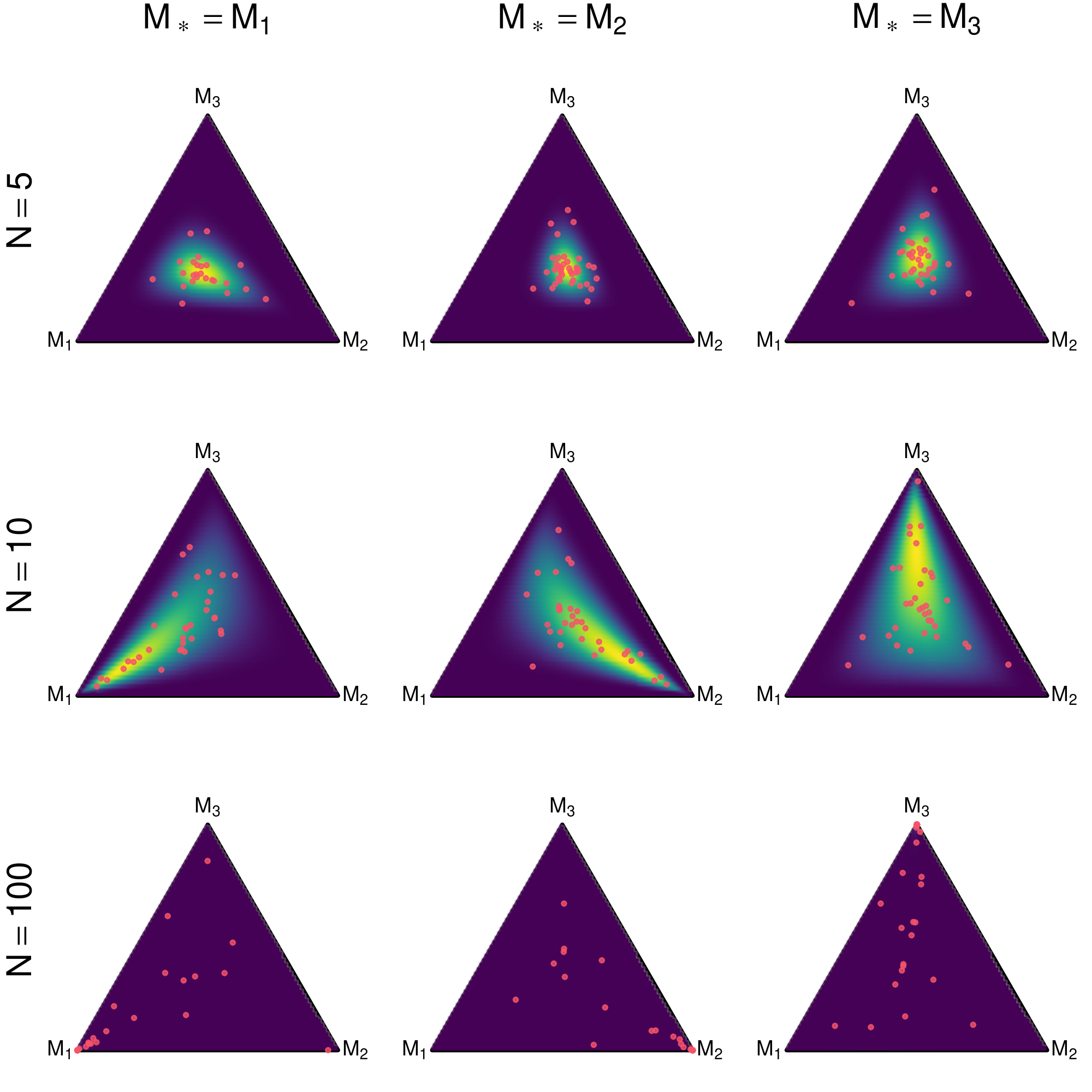}
    \caption{
    \textbf{Experiment 1.} For each $N\in\{5, 10, 100\}$, we simulate $K=100$ data sets and compute the PMPs under each model (red scatter).
    The pushforward likelihood distribution of the expected value for each meta-distribution is illustrated as a density plot in the \texttt{viridis} palette.
    The density in the row $N=100$ is heavily concentrated at the respective vertex $M_*$ and thus hardly visible.}
    \label{fig:experiment-1-level-2-level-3}
\end{figure}

\section{EXPERIMENTS}
In the following, we will illustrate the utility of our uncertainty framework in three BMC scenarios.
For all experiments, we assume uniform model and meta-model priors: $p(M_j) = p(\mathbb{M}_j) = 1/J$. 
Further, we use the default priors of the \texttt{brms} R package for the meta-model parameters and draw $D=2000$ posterior samples from each meta-model.
All meta-models are represented by logistic normal distributions.
The code is publicly available at \url{https://github.com/marvinschmitt/MetaUncertaintyPaper}.

\subsection{Experiment 1: Toy Conjugate Model}
\label{sec:experiments:aleatoric-epistemic}
\paragraph{Setup}
Building on a conjugate Bayesian linear regression model, we aim to characterize the uncertainty at levels 1--3 in a controlled setting.
Thus, we assume three competing regression models~$\M=\{M_1, M_2, M_3\}$, which follow the Normal-Inverse-Gamma conjugacy \autocite{murphy_conjugate_2007},
\begin{equation}\label{eq:experiment-aleatoric-epistemic-model-formulation}
    \begin{aligned}
    \sigma^2 &\sim \Gamma^{-1}(a_0, b_0), \\
    \beta &\sim \mathcal{N}(\mu_0, \sigma^2\Lambda_0^{-1}),\\
    y &\sim \mathcal{N}(\mathbf{X}\beta, \sigma^2 \mathbb{I}),
    \end{aligned}
\end{equation} 
where $\Gamma^{-1}(a, b)$ is the inverse gamma distribution with shape~$a$ and rate~$b$, and $\mathcal{N}(\mu, \Sigma)$ is the multivariate normal distribution with location~$\mu$ and covariance matrix~$\Sigma$.
All models share the prior parameters ${a_0=1}, {b_0=1}, {\mu_0=\0}, {\Lambda_0=\diag(5)}$.
While all models feature the predictor variables $x_1, x_2$, each model has a \emph{distinct} third predictor.

\begin{figure}[t]%
    \centering%
    \begin{subfigure}[t]{0.33\linewidth}%
    \includegraphics[width=\linewidth]{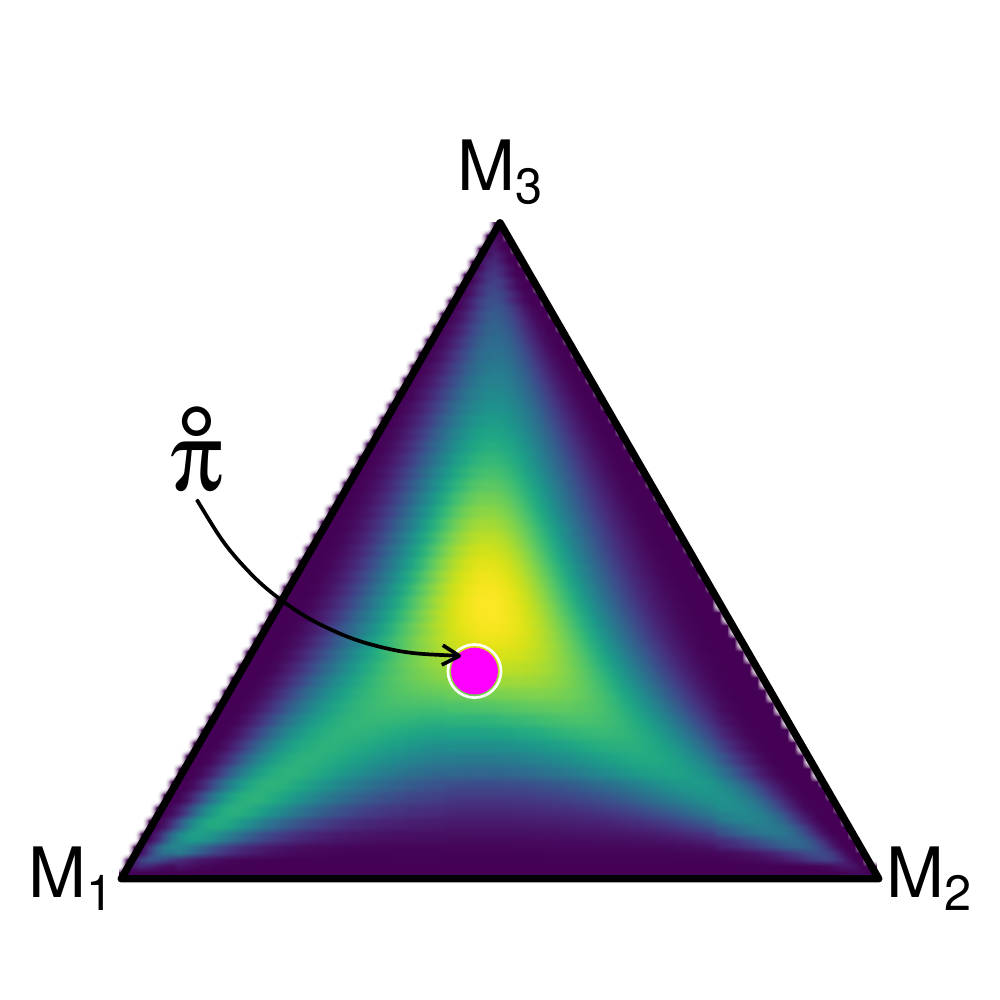}%
    \caption{$\observed{\pi}$\tiny${=}(0.38, 0.31, 0.32)$}%
    \label{fig:experiment-1-predictive-mixture:a}%
    \end{subfigure}%
    \hfill%
    \begin{subfigure}[t]{0.33\linewidth}%
    \includegraphics[width=\linewidth]{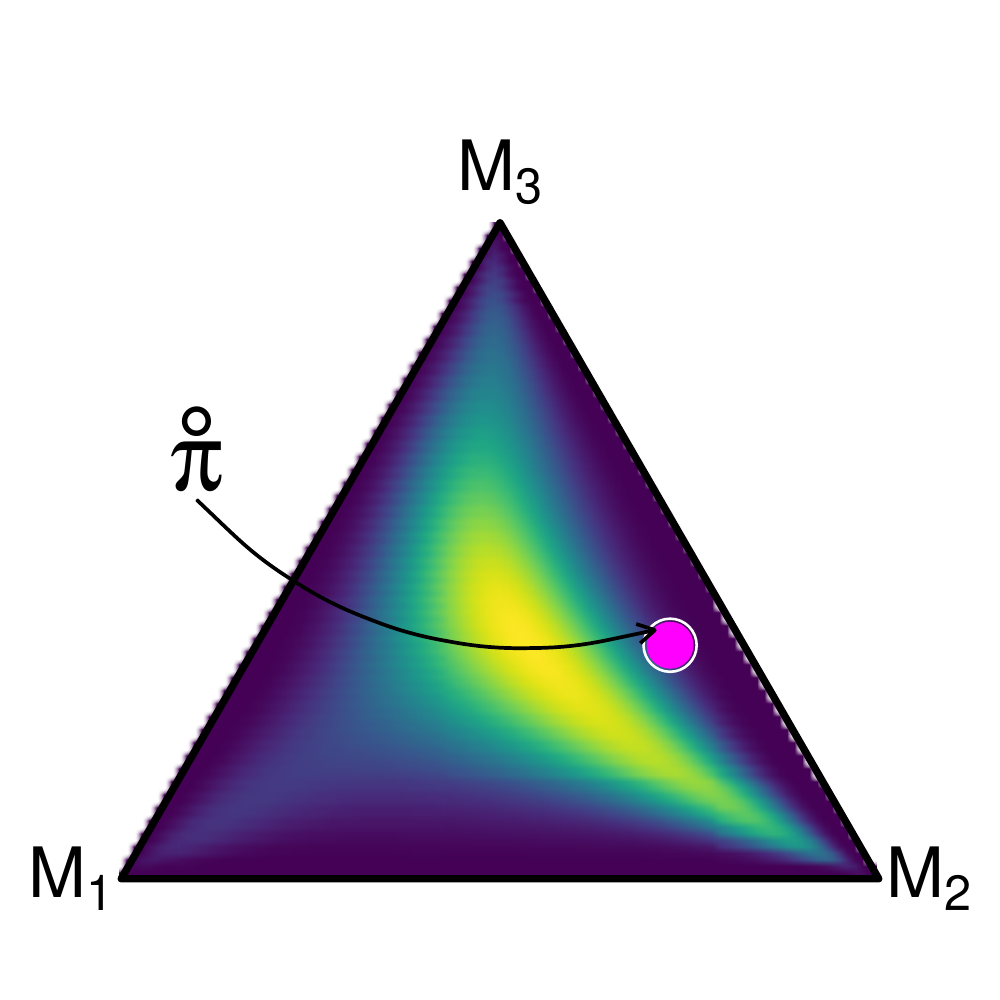}%
    \caption{$\observed{\pi}$\tiny${=}(0.10, 0.55, 0.36)$}%
    \label{fig:experiment-1-predictive-mixture:b}%
    \end{subfigure}%
    \hfill%
    \begin{subfigure}[t]{0.33\linewidth}%
    \includegraphics[width=\linewidth]{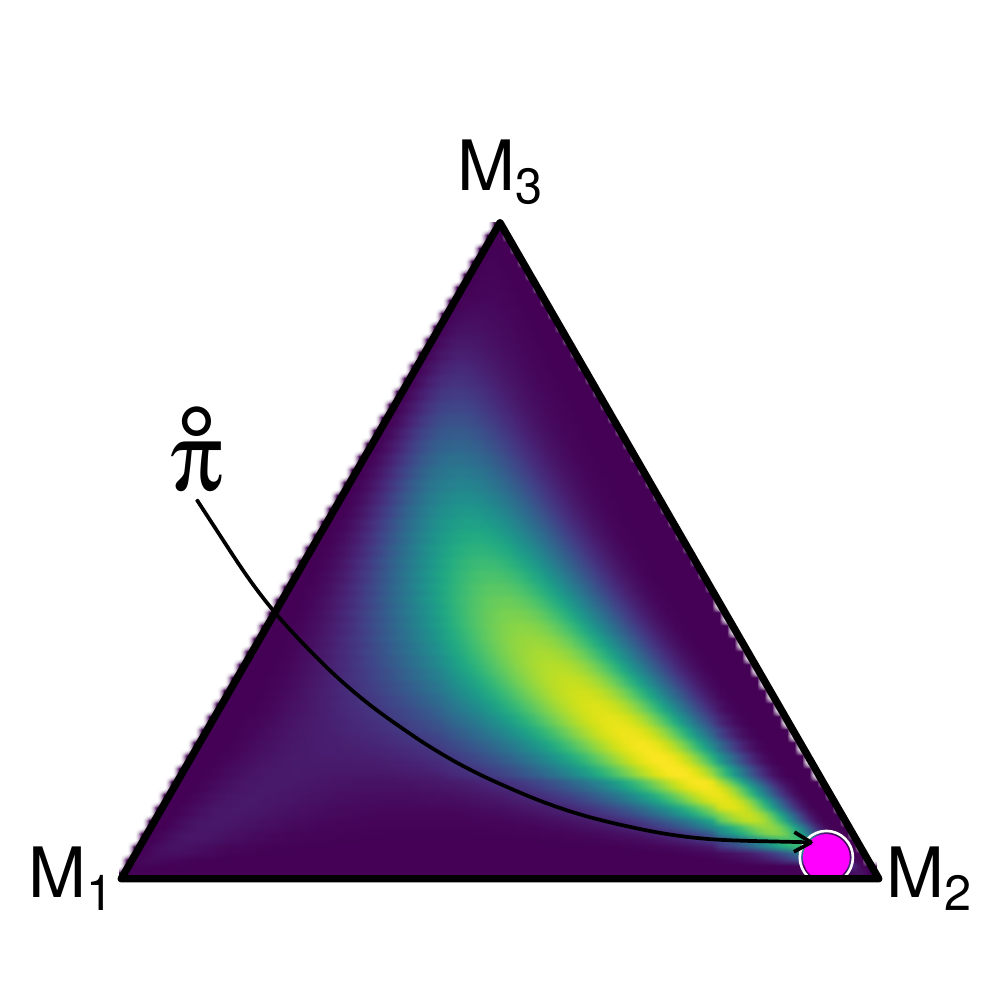}%
    \caption{$\observed{\pi}$\tiny${=}(0.05, 0.91, 0.03)$}%
    \label{fig:experiment-1-predictive-mixture:c}%
    \end{subfigure}%
    \caption{\textbf{Experiment 1.} Predictive mixtures for three \emph{observed} data sets with $N=10$ observations each.}%
    \label{fig:experiment-1-predictive-mixture}%
\end{figure}%

\paragraph{Results} 

As expected, \autoref{fig:experiment-1-level-2-level-3} shows that there is substantial epistemic level 1 uncertainty in the resulting PMPs when the number of observations per data set is small (i.e., $N=5$).
As the number of observations increases, the epistemic uncertainty decreases: The PMPs gravitate towards the true data generating model, that is, the corresponding vertex $M_*$.
The aleatoric variation in the data generating process of $y$ through the nuisance parameter $\Lambda$ propagates into level 2 uncertainty of the corresponding PMPs.
This level 2 uncertainty manifests itself as a scatter across the probability simplex, which corresponds to the \emph{Dance of Bayes Factors} \autocite{oelrich_when_2020}.
In practical applications, this simulation setup can help researchers understand the variation \autocite[i.e., \emph{dance};][]{oelrich_when_2020} of PMPs due to aleatoric uncertainty in the data generating process. 

In order to study meta-uncertainty, we fit meta-models on $K=200$ simulated data sets from the prior predictive distribution for each $N\in\{5, 10, 100\}$.
We aggregate the uncertainty over posterior draws by computing the pushforward likelihood density for the posterior expectation (see \autoref{fig:experiment-1-level-2-level-3}).
As expected, the uncertainty of the pushforward likelihood distribution depends on the uncertainties at level 1 and level 2: For $N=100$, the vanishing level 1 and level 2 uncertainties imply that the posterior predictive distributions concentrate at the corresponding true model $M_*$.

For three \emph{observed} data sets with $N=10$ observations each, the predictive mixture distribution does in fact deviate from the observed PMPs (see \autoref{fig:experiment-1-predictive-mixture}): In (\subref{fig:experiment-1-predictive-mixture:a}) the observed PMPs are uncertain which model is true, so the predictive mixture weighs all three components roughly equally. 
In (\subref{fig:experiment-1-predictive-mixture:b}) the observed PMPs rule out $M_1$ while being balanced between $M_2$ and $M_3$.
This pattern, in turn, manifests itself in the predictive mixture, which is strongly influenced by the components 2 and 3 and shows low density towards $M_1$.
In (\subref{fig:experiment-1-predictive-mixture:c}) the observed PMPs provide strong evidence for $M_2$. 
However, even if $M_2$ was actually true, another researcher gathering new data for a replication study would be likely to find PMPs that provide \emph{much less} evidence for $M_2$, as indicated by the density of the predictive mixture.

\begin{figure*}[t]%
    \centering%
    \begin{subfigure}[t]{0.25\linewidth}%
        \includegraphics[width=\linewidth]{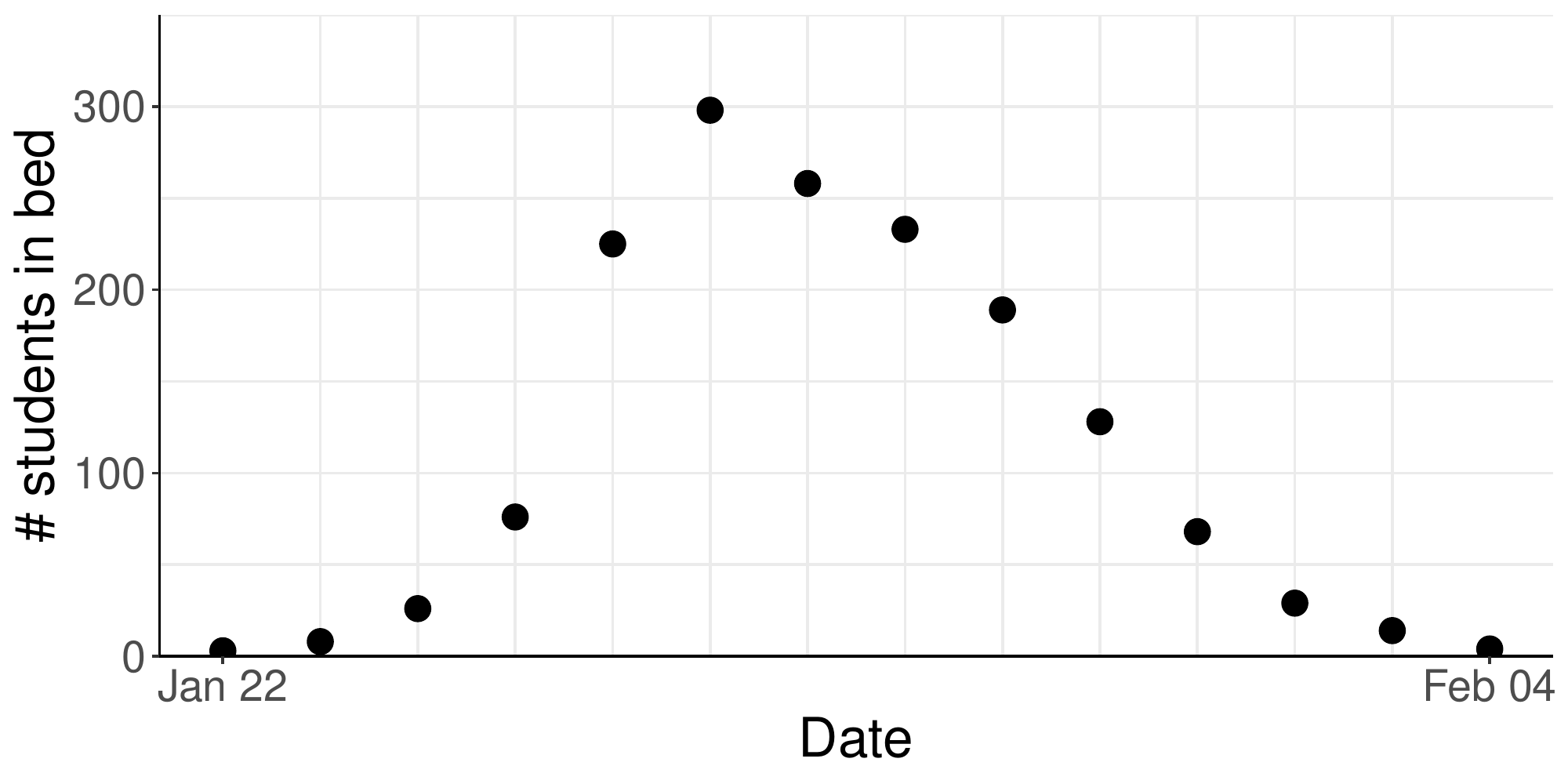}%
        \caption{Observed data set~$\observed{y}$.}%
        \label{fig:experiment-2-data}%
    \end{subfigure}%
    \hfill%
    \begin{subfigure}[t]{0.48\linewidth}%
        \includegraphics[width=\linewidth]{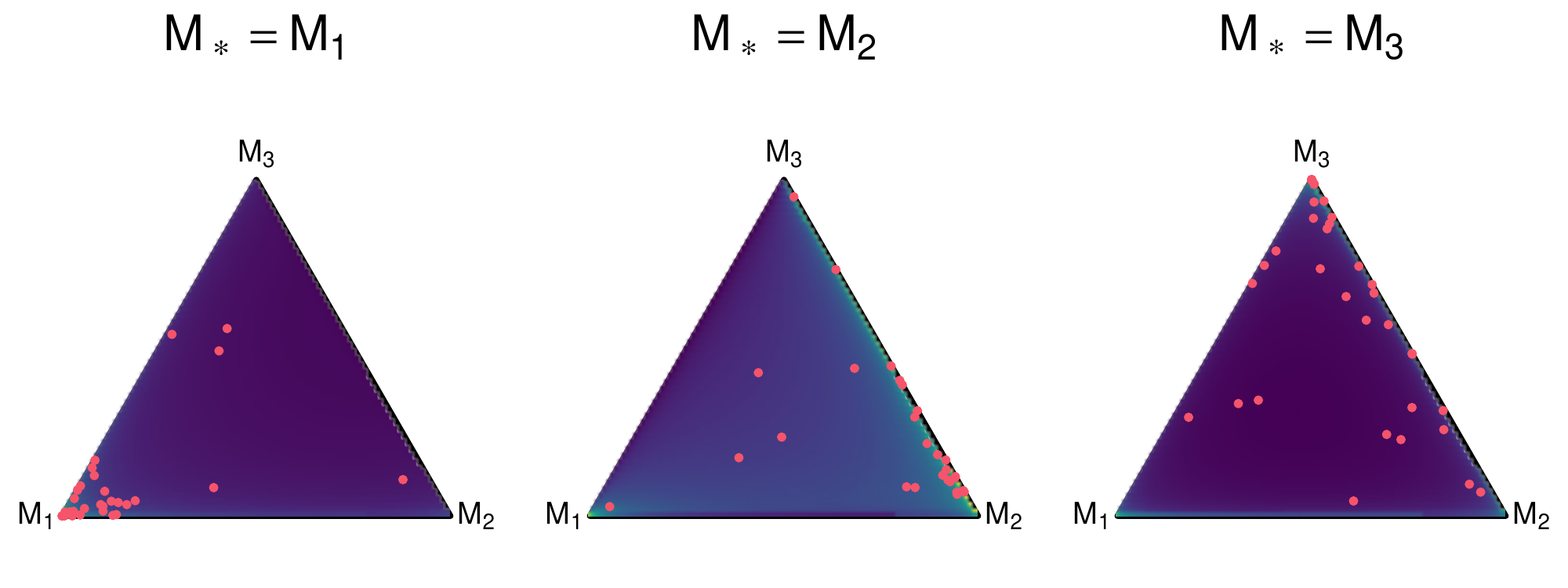}%
        \caption{Level 2 and level 3 uncertainty.}%
        \label{fig:experiment-2-level-2-level-3}%
    \end{subfigure}%
    \hfill%
    \begin{subfigure}[t]{0.16\linewidth}%
        \includegraphics[width=\linewidth, trim={0 -1.8mm 0 15mm},clip]{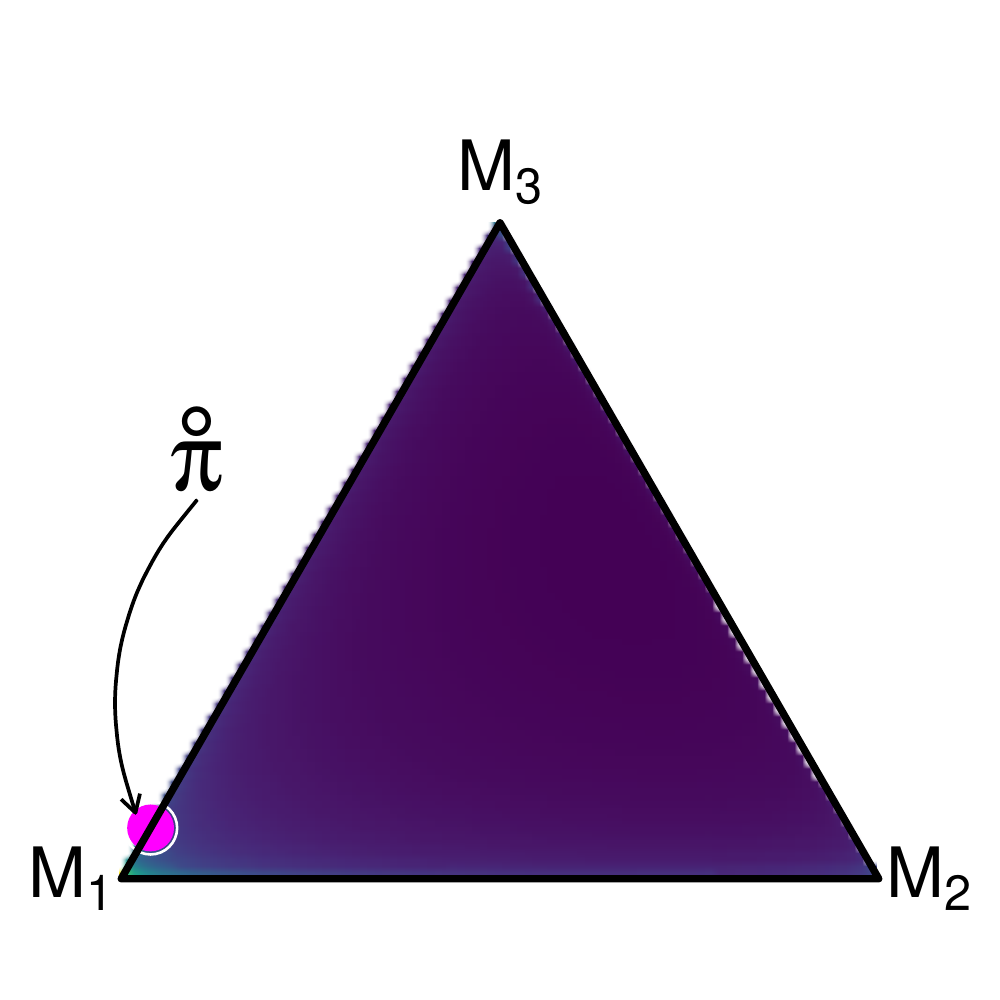}
        \caption{Pred. mixture.}%
        \label{fig:experiment-2-predictive-mixture}%
    \end{subfigure}%
    \caption{\textbf{Experiment 2.} Observed time series~$\observed{y}$, level 2 and level 3 uncertainty, as well as the predictive mixture for the observed time series~$\observed{y}$ of interest.
    All density plots use a $4^{\text{th}}$ root color mapping to the \texttt{viridis} palette to enhance visibility.}%
    \label{fig:experiment-2-results}%
\end{figure*}%

\subsection{Experiment 2: Likelihood-Based Inference}
\paragraph{Setup}
This experiment studies meta-uncertainty in likelihood-based BMC for epidemiological compartmental models of influenza outbreaks using Stan \autocite{stan_development_team_stan_2022} and bridge sampling \autocite{gronau_bridgesampling_2020}. The experiment and software code are adapted from \textcite{grinsztajn_bayesian_2021}.
The three assumed models $M_1, M_2, M_3$ differ with respect to their latent disease model (SIR vs.\ SEIR) and their likelihood (Poisson vs.\ Negative-Binomial).
In this experiment, $M_1$ is the least complex model (SIR, Poisson), followed by $M_2$ (SEIR, Poisson), and the most complex model $M_3$ (SEIR, Negative-Binomial).
The \textbf{Supplementary Material} contains implementation details and model configurations.

The observed data set~$\observed{y}$ stems from an influenza (H1N1) outbreak in 1978 at a British boarding school with ${N=763}$ students, available in the \texttt{outbreaks} R package \autocite{jombart_outbreaks_2020}.
The data set contains the number of students in bed $y$ at each day from January 22 to February 4, see \autoref{fig:experiment-2-data}.

\paragraph{Results} The predictive mixture method on $K=100$ simulated data sets illustrates that the model-implied PMPs in \autoref{fig:experiment-2-level-2-level-3} are aligned with the strong observed PMPs~${\observed{\pi}=(0.92, {<}0.0001, 0.08)}$, which supports the conclusion that the latter are \emph{not} overconfident.
Accordingly, the predictive mixture distribution is concentrated at the observed PMPs, too (see \autoref{fig:experiment-2-predictive-mixture}), and we would expect a replication study to show equally strong evidence for $M_1$.

\subsection{Experiment 3: Simulation-Based Inference}\label{sec:experiments:amortized-pmp}

\paragraph{Setup}
This experiment will demonstrate that the predictive mixture method by itself is essentially independent of the complexity of the data generating process: 
Since it acts directly on PMPs and these PMPs are agnostic to the employed approximation method, the difficulty of calculating PMPs does not affect our predictive mixture method.
Accordingly, we will compare three Hodgkin-Huxley simulator models of neural activity~$\M=\{M_1, M_2, M_3\}$ with different parameter configurations~$\Theta_{\mathcal{M}} = \{\theta_1, \theta_2, \theta_3\}$ \autocite{hodgkin_quantitative_1952, pospischil_minimal_2008} using the BayesFlow methodology~\autocite{radev_amortized_2021}.



Similarly to \textcite{radev_amortized_2021}, we train an unregularized evidential network for 25 epochs with 500 iterations (i.e., batch simulations) per epoch. 
Training details and loss history are described in the \textbf{Supplementary Material}.
In the inference phase, the network estimates the parameters of an approximate Dirichlet distribution~$f_{\phi}(M\,|\,y)$ over the PMPs for each simulated neural time series~$y$.
Here, $\phi$ denotes all parameters of the trained network.
We derive a point estimate~$\hat{p}(M\given y)$ from the expected value of the Dirichlet density as~$\mathbb{E}_{\pi\sim\mathrm{Dir}(\alpha)}[\pi_j]=\alpha_j / \sum_j\alpha_{j}$ \autocite{balakrishnan_primer_2003}.

Within our meta-uncertainty framework, we use the trained network to perform fast amortized model comparison on $K=200$ simulated neural time series, effectively leveraging the network's ability to calculate the PMPs in line~\ref{alg:model-implied-pmp-distribution:pmp-calculation} of \autoref{alg:model-implied-pmp-distribution} in a fraction of a second.
The fast amortized inference of PMPs enables us to use the predictive mixture method on remarkably complex models, since it bypasses the bottleneck of fitting a large number of models ($J$ models for each of the $K$ data sets) to obtain the required set of model-implied PMPs.
This renders the second phase of our method (i.e., fitting the meta-models) essentially independent of the complexity of the data generating process.

\paragraph{Results} 
\begin{figure}[t]
    \centering
    \includegraphics[width=\linewidth]{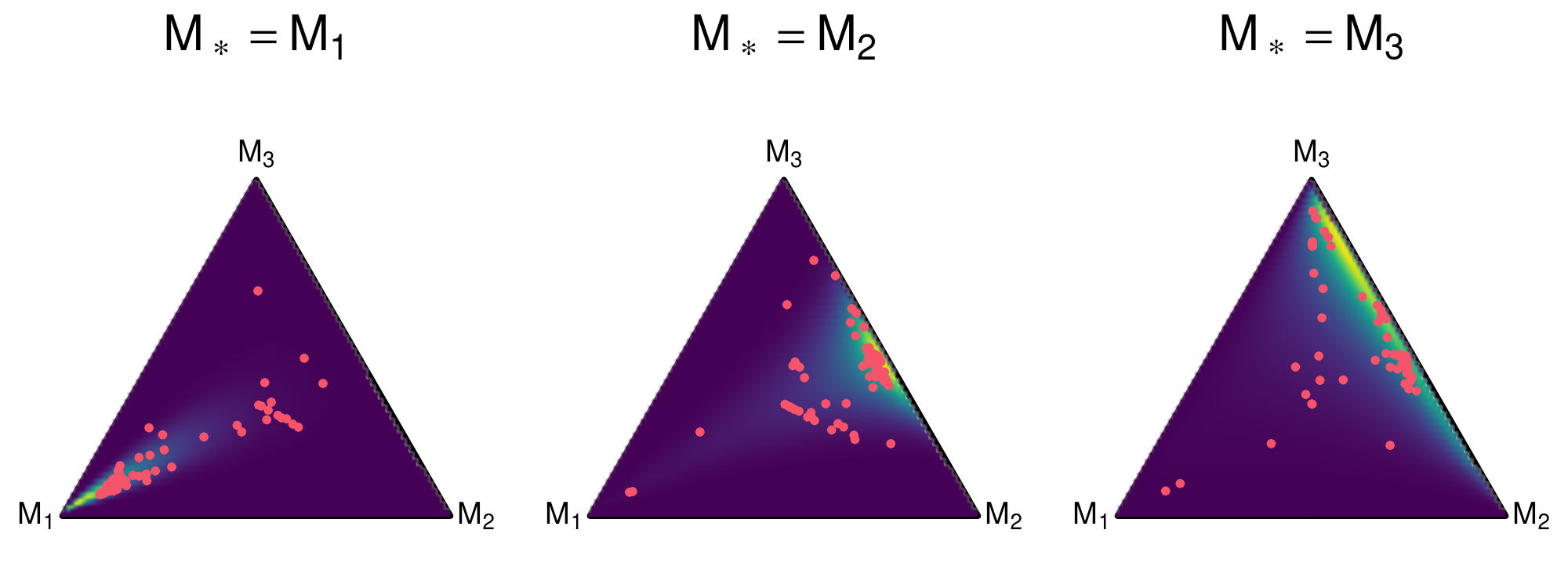}
    \caption{\textbf{Experiment 3.} Model-implied distributions of PMPs on $K=200$ simulated time series (level 2) are depicted as scatter across the probability simplices.
    The pushforward likelihood distributions for the meta-models' expected posterior values are illustrated in the viridis palette.}
    \label{fig:experiment-3-level-2-level-3}
\end{figure}
The model-implied PMP distributions (level 2) as well as the pushforward likelihoods for the expectations of the meta-models (level 3) are depicted in \autoref{fig:experiment-3-level-2-level-3}.
There is substantial level 2 uncertainty across all data generating models.
However, it is worth noting that data from $M_2$ rarely ever features high probability for $M_2$ itself.
In contrast, a large fraction of data sets simulated by $M_1$ attains high posterior model probability for $M_1$.
The estimated meta-models are able to reveal these patterns.

\begin{figure}[t]
    \centering
    \begin{subfigure}[t]{0.32\linewidth}
        \includegraphics[width=\linewidth]{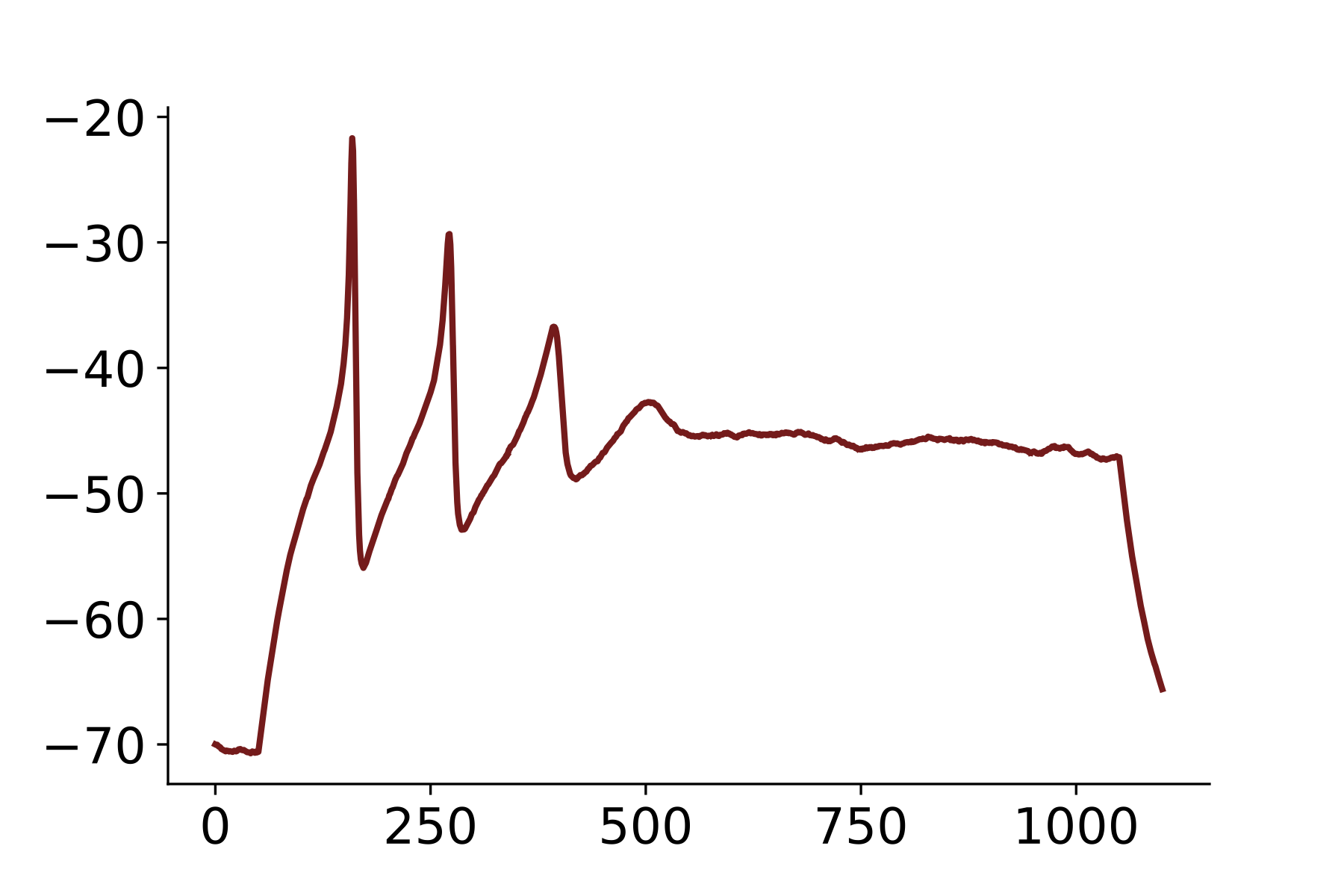}
        \caption{Observed data~$\observed{y}$.}
        \label{fig:experiment-3-obs}
    \end{subfigure}
    \hfill
    \begin{subfigure}[t]{0.32\linewidth}
        \includegraphics[width=\linewidth]{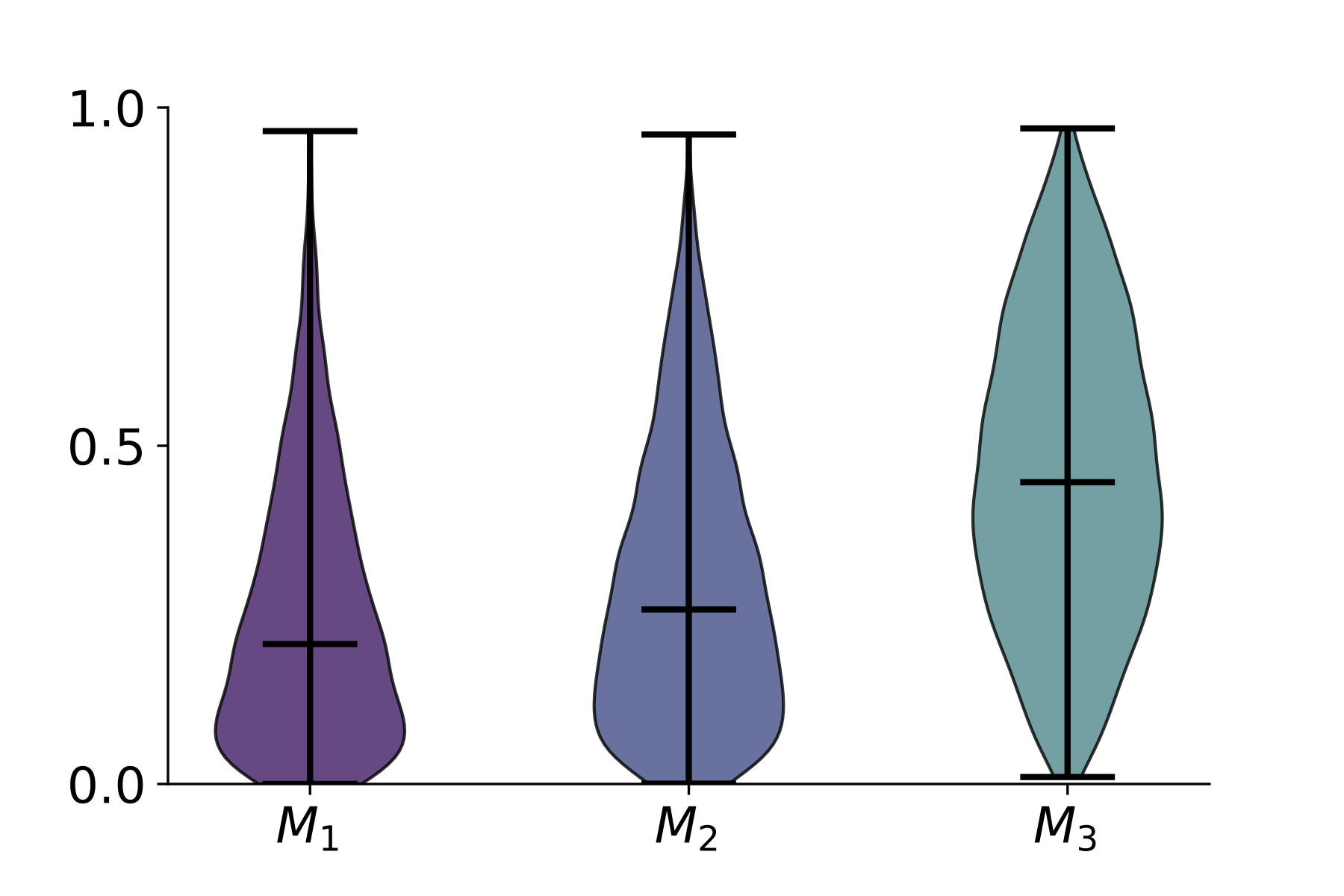}
        \caption{Estimated PMPs $\observed{\pi}$ via BayesFlow.}
    \end{subfigure}
    \hfill
    \begin{subfigure}[t]{0.32\linewidth}
        \includegraphics[width=\linewidth]{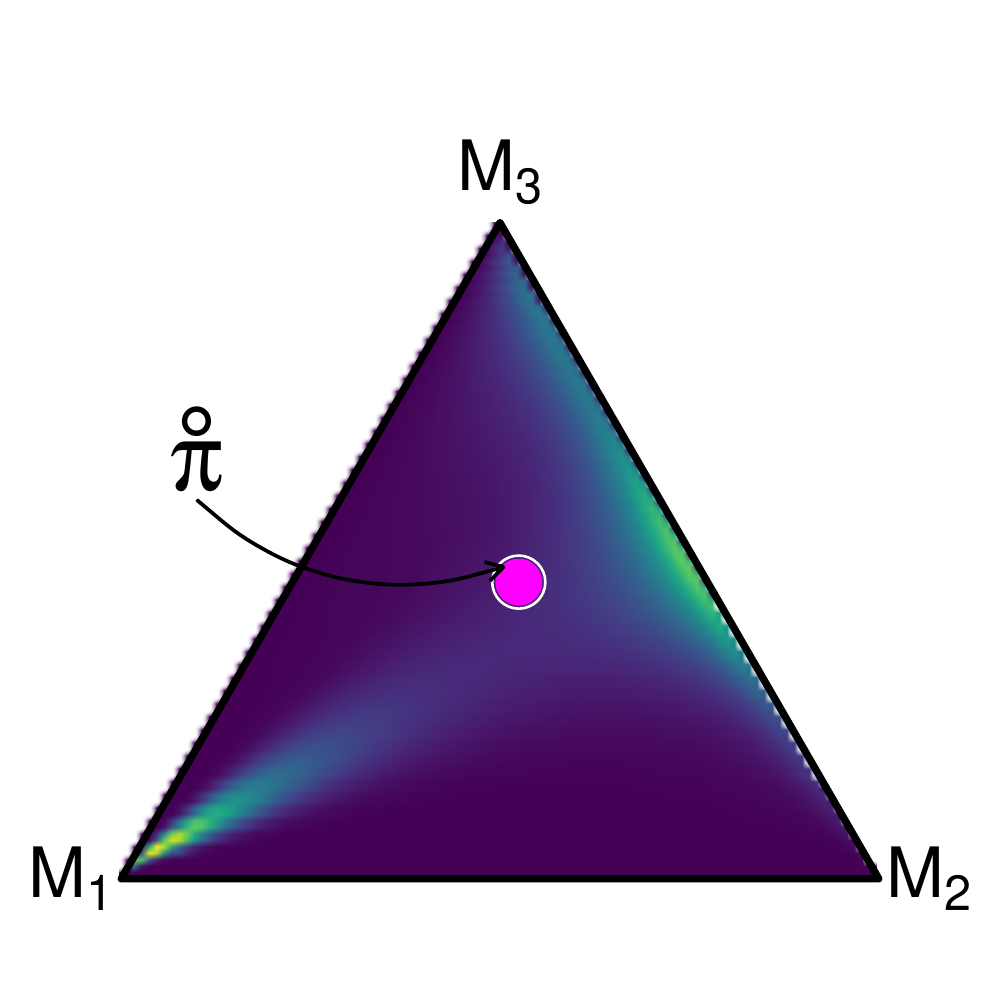}
        \caption{Predictive mixture distribution.}
        \label{fig:experiment-3-predictive-mixture-predictive-mixture-distribution}
    \end{subfigure}
    \caption{\textbf{Experiment 3.} For the observed data set $\observed{y}$, BayesFlow yields an estimated PMP distribution.
    Our predictive mixture augments this estimate with model-implied uncertainty, revealing the multi-model distribution of predicted PMPs for new data.}
    \label{fig:experiment-3-predictive-mixture-figure}
\end{figure}

Finally, approximating the PMPs for an observed neural time series~$\observed{y}$ (see \autoref{fig:experiment-3-obs}) yields $\observed{\pi}{=}(0.25, 0.30, 0.45)$.
Our predictive mixture method equips the observed PMPs with information from realizations of the complex data generating process.
Correspondingly, \autoref{fig:experiment-3-predictive-mixture-figure} shows that the predictive mixture distribution deviates from the observed PMPs estimated using BayesFlow.
While the observed PMPs suggest a slight tendency towards $M_3$, the simulation-informed predictive mixture distribution unveils the strongly multimodal structure of the predicted PMPs for a new data set:
Areas of high density in \autoref{fig:experiment-3-predictive-mixture-predictive-mixture-distribution} anticipate either a large PMP for $M_1$, or else balanced PMPs for $M_2$ and $M_3$, with low posterior probability for $M_1$.



\section{CONCLUSION}
We have proposed a method to extend the methodological repertoire of probabilistic modeling workflows.
Our method does not purport to provide a ``better'' point estimate than standard posterior model probabilities or Bayes factors.
Using the generative property of Bayesian models, our \emph{predictive mixture} method merely \emph{augments} posterior model probabilities with simulation-based information.
This information yields a more nuanced picture of uncertainty in finite-data settings and can prevent overconfident model selection in practice.

Meta-uncertainty can readily be derived for any (standard or adjusted) PMPs by replacing the marginal likelihood in line~\ref{alg:model-implied-pmp-distribution:pmp-calculation} of \autoref{alg:model-implied-pmp-distribution}, for example by conditional \autocite[aka.\ partial;][]{ohagan_fractional_1995, lotfi_bayesian_2022}, fractional \autocite{ohagan_fractional_1995}, or outlier-adjusted marginal likelihoods.

Future work could systematically investigate the connections between meta-models and related concepts from evidence theory \autocite{shafer_mathematical_1976, deng_generalized_2015}.
In addition, the consequences of using different meta-models (i.e., probability distributions over probabilities) for the predictive mixture distribution can be further explored.

We hope that our framework can reveal intriguing empirical properties of Bayesian model probabilities and guide applied model building and model comparison workflows in virtually any research area.
A selection of frequently asked questions we encountered prior to publication is answered in the FAQ section of the \textbf{Supplementary Material}.


\subsubsection*{Acknowledgments}
We thank Lasse Elsemüller for his valuable feedback.
We thank all reviewers for their constructive and thought-provoking feedback, which considerably improved the quality and clarity of the paper.
MS was supported by the Cyber Valley Research Fund (grant number: CyVy-RF-2021-16).
MS and PCB were supported by the Deutsche Forschungsgemeinschaft (DFG, German Research Foundation) under Germany’s Excellence Strategy -- EXC-2075 - 390740016 (the Stuttgart Cluster of Excellence SimTech).
STR was supported by the Deutsche Forschungsgemeinschaft (DFG, German Research Foundation) under Germany’s Excellence Strategy -– EXC-2181 - 390900948 (the Heidelberg Cluster of Excellence STRUCTURES).

\FloatBarrier
\subsubsection*{References}
\printbibliography[heading=none]

\appendix
\onecolumn

\section{Frequently Asked Questions (FAQ)}
\subsubsection*{How can I reproduce the results?}

The full software code (R and Python) to reproduce the minimal example in the Methods section, as well as all experiments, is available in the public repository at \url{https://github.com/marvinschmitt/MetaUncertaintyPaper}.
The code structure is detailed in Section~\ref{sec:app:code} in the Appendix.
\subsubsection*{Is meta-uncertainty better than standard posterior model probabilities?}
We do not intend to provide a ''better`` point estimate than standard posterior model probabilities or Bayes Factors.
Instead, we \emph{enrich} observed PMPs by embedding them into the probabilistic context of the compared models to enable more realistic interpretations in practice.
The epistemic uncertainty in PMPs vanishes for the infinite data case $N{\to}\infty$, and this is a desired property known as consistency.
However, in the \emph{finite data} case, we might encounter \emph{misleadingly} extreme preference for one model, i.e., overconfidence.
The paramount importance of overconfidence is well-known and acknowledged in the scientific community (cf.~Oelrich et al., 2020; Yang \& Zhu, 2018).
A replication study might come to an utterly different result, but both observed results are compatible with the models.
Yet, we cannot learn this from any single study, but it is clearly visible through the level 2 uncertainty of PMPs across many simulated data sets.
Thus, our method mitigates overconfidence by answering the question ''which PMPs do we expect in a replication study{?}`` with a principled Bayesian approach.
In all three experiments, we show how our method yields additional insights beyond single observed PMPs.

\subsubsection*{How is meta-uncertainty related to modified marginal likelihoods?}

The idea of conditional marginal likelihoods to use parts of the data to pre-condition marginal likelihoods reaches back to \textcite{ohagan_fractional_1995}, and it is not an alternative to our method.
In fact, our pipeline readily works for modified computations of PMPs by adjusting line 5 in Algorithm 1, for example through conditional (partial) marginal likelihoods, fractional marginal likelihoods, or outlier-adjusted marginal likelihoods.

\subsubsection*{How is meta-uncertainty related to Bayesian model averaging?}

Bayesian model averaging (BMA) is concerned with parameter estimation under several competing models, and thus not an alternative to our method which is exclusively concerned with model comparison. However, our predictive mixture distribution might serve as a basis for averaged parameter estimates (replacing PMPs in Bayesian model averaging), and this might be a promising future research endeavor.

\subsubsection*{How is meta-uncertainty related to the Bayesian information criterion?}
The Bayesian information criterion (BIC) is neither fully Bayesian, nor an information criterion in a strict sense. Instead, it is a (possibly crude) approximation of the marginal likelihood, which lies at the core of our method

\subsubsection*{How does the uncertainty change with more samples?}
As the number of observations per data set grows arbitrarily large ($N\to\infty$), the epistemic uncertainty in the PMP vectors vanishes. For an increasing number of simulated data sets ($K\to\infty$), we argue that (i) the level 2 distribution of PMPs across data sets approaches the sampling distribution (cf. Section 3.1, “Relation to Sampling Distributions”); and (ii) the epistemic uncertainty of the meta models vanishes (cf. Section 3.2, “Level 3 Uncertainty”).
Furthermore, the levels of the framework are briefly outlined in \autoref{tab:overview}.

\subsubsection*{How do outliers in the data influence the results?}
We acknowledge that outliers in the data~$y$ can lead to overconfident PMPs for single data sets.
Yet, if outliers only occur occasionally in few data sets, the level 2 distribution should mirror this behavior, and the meta model should reflect it too.
Thus, our method naturally puts data sets with outliers into an appropriate probabilistic context of the models without altering the data~$y$ by removing said outliers.

\subsubsection*{Does the method work for models with high-dimensional parameter spaces?}

We agree that settings with high parameter dimensions constitute a paramount application for (Bayesian) model selection, and established Bayesian methods (ABC, MCMC) are infeasible in high dimensions with large data sets.
As discussed in Section 5.3, our method is essentially independent of the data dimensionality, and the method to obtain PMPs (line 5 in Algorithm 1) is the actual bottleneck.
Recent advances in simulation-based inference (SBI) allow model comparison in high dimensions through special neural network architectures, such as BayesFlow \autocite{radev_bayesflow_2020}.
As demonstrated in Experiment 3, our method is fully compatible with these state-of-the-art neural approaches and thus capable of augmenting model comparison workflows in high dimensions.
Furthermore, Experiment 3 shows that our method has a negligible computational cost when an amortized PMP approximator (such as BayesFlow) is available, which further facilitates the integration of our methods to simulation-based inference workflows.

\subsubsection*{Does the method work for large-scale Bayesian variable selection?}
The current state-of-the-art method for large-scale Bayesian variable selection is projection predictive inference, while using PMPs for this purpose is not advisable \autocite{piironen_projective_2020}. The most important application of PMPs, and thus also of our method, are comparisons of qualitatively different models arising from competing scientific theories.
While the parameter spaces may be high-dimensional (and simulation-based inference might be employed), the model space is usually low-dimensional (i.e., only few models are being compared).

\section{CODE}\label{sec:app:code}
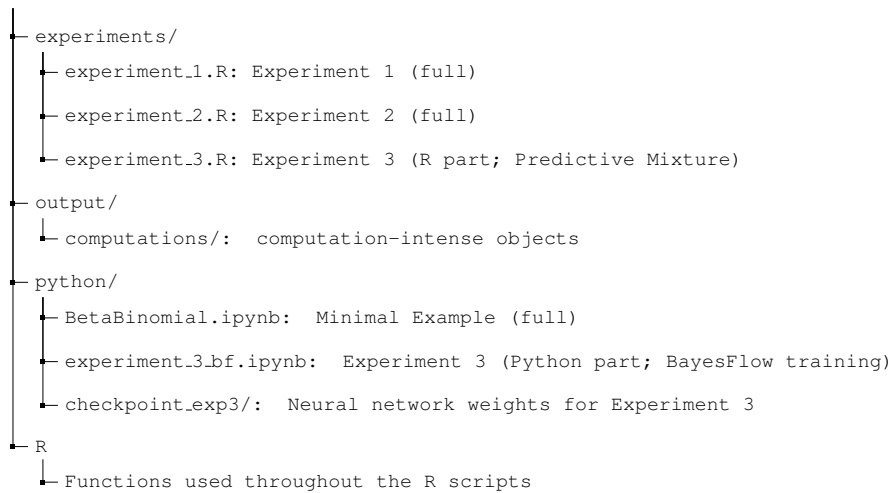
\begin{figure}[h]
    \centering
    \begin{adjustbox}{max width=0.70\textwidth}
\begin{forest}
  for tree={
    font=\ttfamily,
    grow'=0,
    child anchor=west,
    parent anchor=south,
    anchor=west,
    calign=first,
    edge path={
      \noexpand\path [draw, \forestoption{edge}]
      (!u.south west) +(7.5pt,0) |- node[fill,inner sep=1.25pt] {} (.child anchor)\forestoption{edge label};
    },
    before typesetting nodes={
      if n=1
        {insert before={[,phantom]}}
        {}
    },
    fit=band,
    before computing xy={l=15pt},
  }
  [
    [experiments/
        [experiment\_1.R: Experiment 1 (full)]
        [experiment\_2.R: Experiment 2 (full)]
        [experiment\_3.R: Experiment 3 (R part; Predictive Mixture)]
    ]
    [output/
        [computations/: computation-intense objects]
    ]
    [python/
        [BetaBinomial.ipynb: Minimal Example (full)]
        [experiment\_3\_bf.ipynb: Experiment 3 (Python part; BayesFlow training)]
        [checkpoint\_exp3/: Neural network weights for Experiment 3]
    ]
  [R
  [Functions used throughout the R scripts]
  ]
  ]
\end{forest}
\end{adjustbox}
    \caption{Outline of the experimental code.}
    \label{fig:code-outline}
\end{figure}

Code for all methods and experiments is available in the public repository at \url{https://github.com/marvinschmitt/MetaUncertaintyPaper}.
The code is structured as outlined in \autoref{fig:code-outline}.
To facilitate reproducibility, we use seeds for the random number generators throughout the analyses.
Note that the \texttt{R} code is structured as a package, thus requiring a local installation with subsequent loading via \texttt{library(MetaUncertaintyPaper)}.

\clearpage
\section{EXPERIMENTS: IMPLEMENTATION DETAILS AND ADDITIONAL RESULTS}
\subsection{Experiment 1: Conjugate Model}
The conjugate construction yields an analytic posterior distribution for the inference parameters \autocite{murphy_conjugate_2007}, which again follow a Normal-Inverse-Gamma distribution $\mathcal{N}\text{-}\Gamma^{-1}(a_n, b_n, \mu_n, \Lambda_n)$ with
\begin{equation}
	\begin{aligned}
		a_n & = a_0 + \frac{n}{2},\\
		b_n & = b_0 + \frac{1}{2} (y^\top y + \mu_0^\top\Lambda_0\mu_0 - \mu_n^\top\Lambda_n\mu_n),\\
		\mu_n & = (\X^\top\X + \Lambda_0)^{-1}(\X^\top\X(\X^\top\X)^{-1}\X^\top y+\Lambda_0\mu_0)\\
		\Lambda_n & = \X^\top\X + \Lambda_0.
	\end{aligned}
\end{equation}
Consequently, the marginal likelihood can be directly calculated \autocite{murphy_conjugate_2007, ohagan_kendalls_1994} to obtain the analytic expression
\begin{equation}\label{eq:experiment-2-j6}
	\begin{aligned}
		p(y\given M) = 
		\dfrac{1}{(2\pi)^{n/2}}\,
		\sqrt{\dfrac{\det(\Lambda_0)}{\det(\Lambda_n)}}\,
		\dfrac{b_0^{a_0}}{b_n^{a_n}}\,
		\dfrac{\Gamma(a_n)}{\Gamma(a_0)}.
	\end{aligned}
\end{equation}

\begin{figure}
    \centering
    \includegraphics[width=0.60\linewidth]{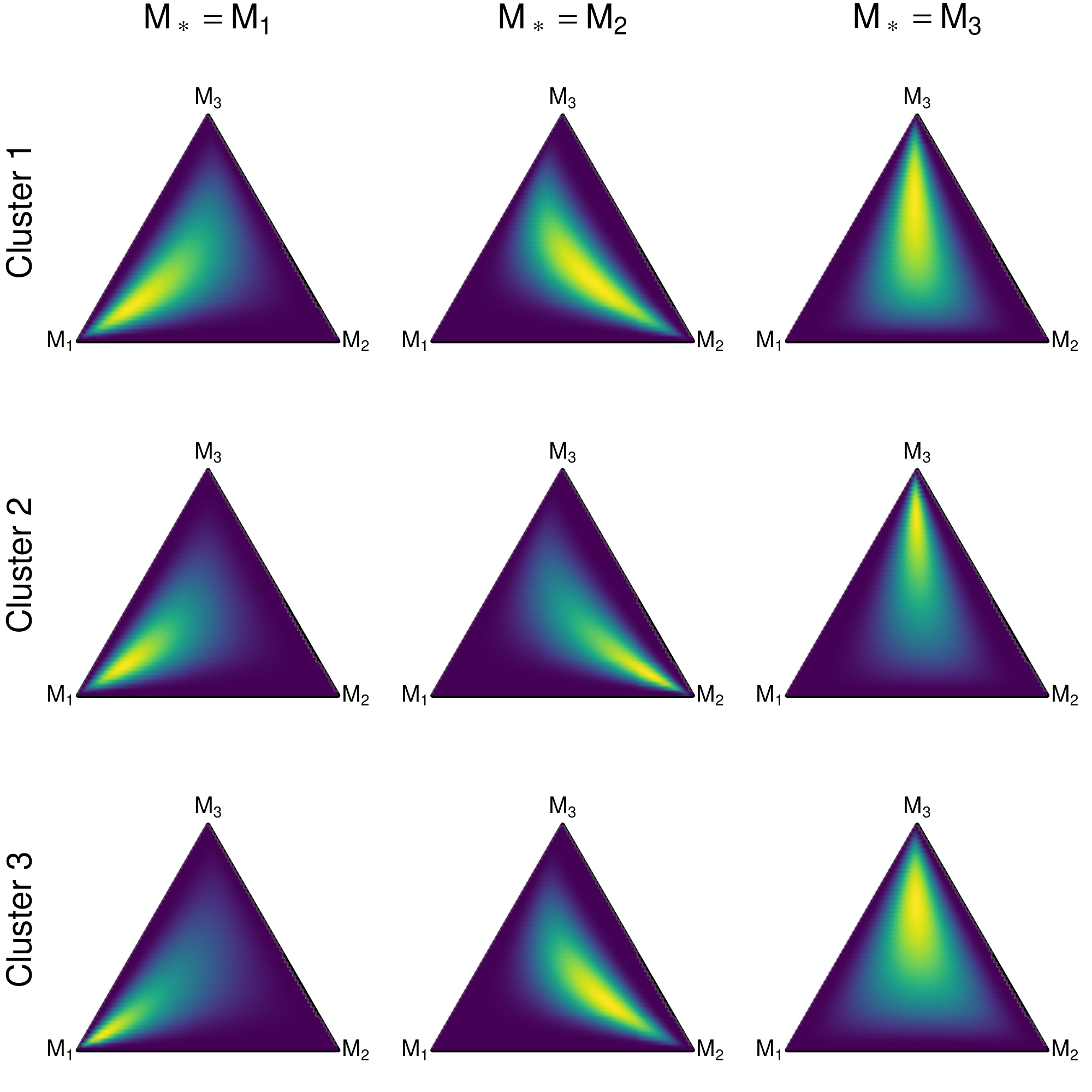}
    \caption{\textbf{Experiment 1.} Pushforward densities of cluster means from the posterior draws of the meta models on simulated data.}
    \label{fig:experiment-1-postpred-clusters}
\end{figure}


\FloatBarrier
\subsection{Experiment 2: Likelihood-based Inference}
Following the implementation of \textcite{grinsztajn_bayesian_2021}, the data generating process for the influenza model is formulated either as an SIR model or as an SEIR model
\begin{equation}
\begin{aligned}
        \text{SIR:} \qquad& 
        &\frac{\diff S}{\diff t} &= -\beta S\frac{I}{N},\qquad
        & &
        &\frac{\diff I}{\diff t} &= \beta S\frac{I}{N}-\gamma I,\qquad
        &\frac{\diff R}{\diff t} &= \gamma I, \\
        \text{SEIR:} \qquad& 
        &\frac{\diff S}{\diff t} &= -\beta S\frac{I}{N},\qquad
        &\frac{\diff E}{\diff t} &= \beta S\frac{I}{N} - \eta E,\qquad
        &\frac{\diff I}{\diff t} &= \eta E-\gamma I,\qquad
        &\frac{\diff R}{\diff t} &= \gamma I,
\end{aligned}
\end{equation}
with susceptible ($S$), exposed ($E$), infected ($I$), and recovered ($R$) population compartments, respectively.
The likelihood function represents the stochastic relationship between the latent counts in the $I$ compartment and the actually reported infections (i.e., students in bed) $y$ as either a Negative-Binomial distribution
\begin{equation}
    p(y\given \theta) = \text{Negative-Binomial}(y\given I(t), \phi)
\end{equation}
or as a Poisson distribution
\begin{equation}
    p(y\given \theta) = \mathrm{Poisson}(y\given I(t))
\end{equation}

where $\theta$ expresses all parameters in the respective model, e.g., $\theta=\{\beta, \gamma, \eta, \phi\}$ for a latent SEIR model with Negative-Binomial likelihood.
The prior distributions are specified as

\begin{equation}
    \begin{aligned}
        \beta  \sim \N^+(\mu_{\beta}, \sigma_{\beta}),\qquad
        \gamma \sim \N^+(\mu_{\gamma}, \sigma_{\gamma}),\qquad
        \eta \sim \N^+(\mu_{\eta}, \sigma_{\eta}), \qquad
        \phi \sim \N^+(\mu_{\phi}, \sigma_{\phi}),
    \end{aligned}
\end{equation}

where $\N^{+}$ denotes the (positive) half-Normal distribution truncated at 0.
The model configurations are outlined in \autoref{tab:experiment-2-model-priors}.

\begin{table}[b]
	\caption{Competing models in \textbf{Experiment 2}. The models differ with respect to their latent epidemiological compartmental model (CM) and the functional form of the likelihood.}
	\label{tab:experiment-2-model-priors}
	\centering
	\begin{tabular}{r|cccccccccc}
		& Latent CM & Likelihood &$\mu_{\beta}$ & $\sigma_{\beta}$ & $\mu_{\gamma}$ & $\sigma_{\gamma}$ & $\mu_{\eta}$ & $\sigma_{\eta}$ & $\mu_{\phi}$ & $\sigma_{\phi}$\\
		\hline 
		$M_1$ & SIR  & Poisson           & $2$ & $0.1$ & $0.4$ & $0.1$   & --- & ---   & --- & --- \\
		$M_2$ & SEIR & Poisson           & $3$ & $0.1$ & $0.5$ & $0.1$   & $3$ & $0.1$   & --- & ---\\
		$M_3$ & SEIR & Negative-Binomial & $3$ & $0.1$ & $0.5$ & $0.1$   & $3$ & $0.1$   & 100 & 1
	\end{tabular}
\end{table} 


\begin{figure}[t]
    \centering
    \begin{subfigure}[t]{0.32\linewidth}
        \includegraphics[width=\linewidth]{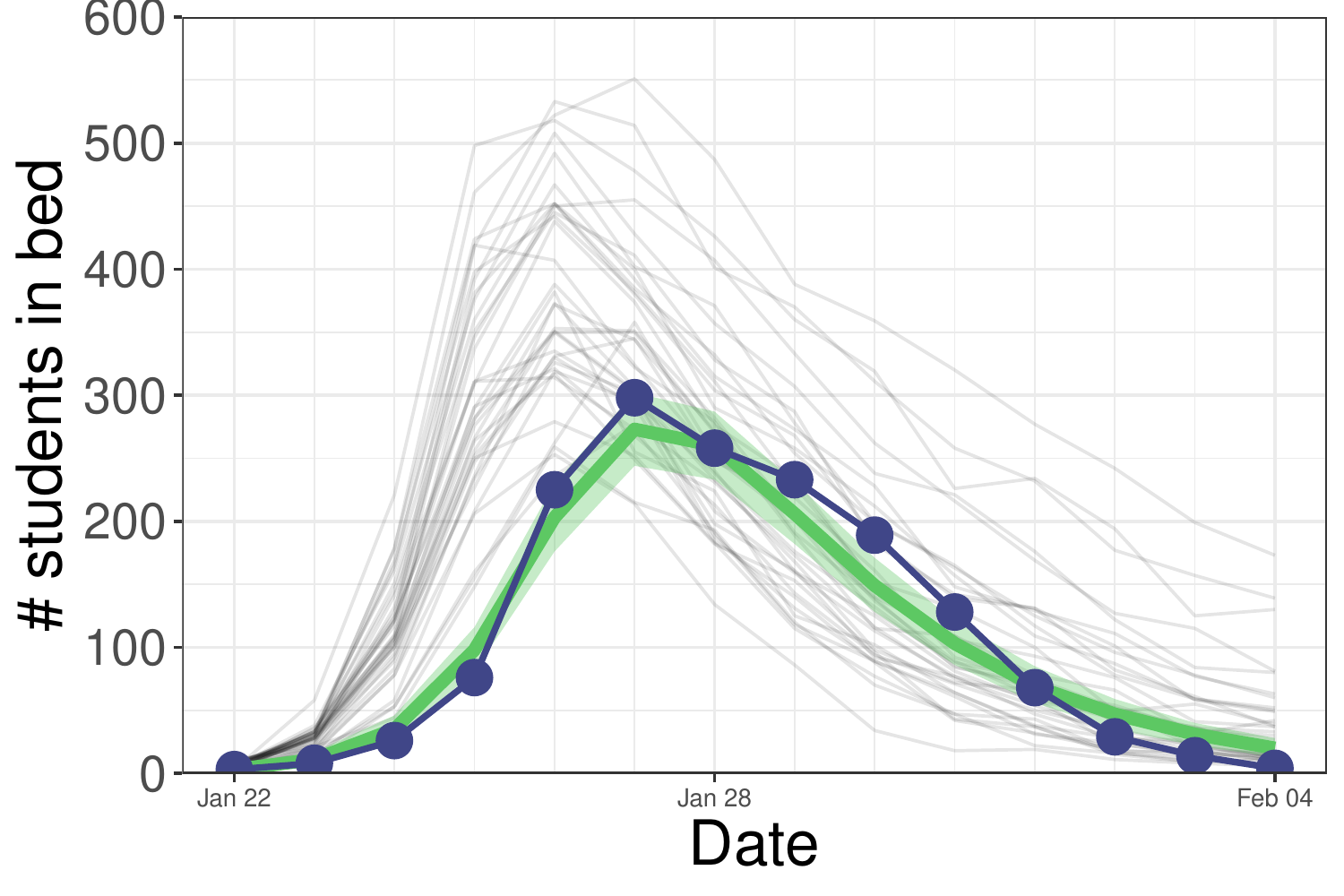}
        \caption{$M_1$: SIR, Poisson}
    \end{subfigure}
    \begin{subfigure}[t]{0.32\linewidth}
        \includegraphics[width=\linewidth]{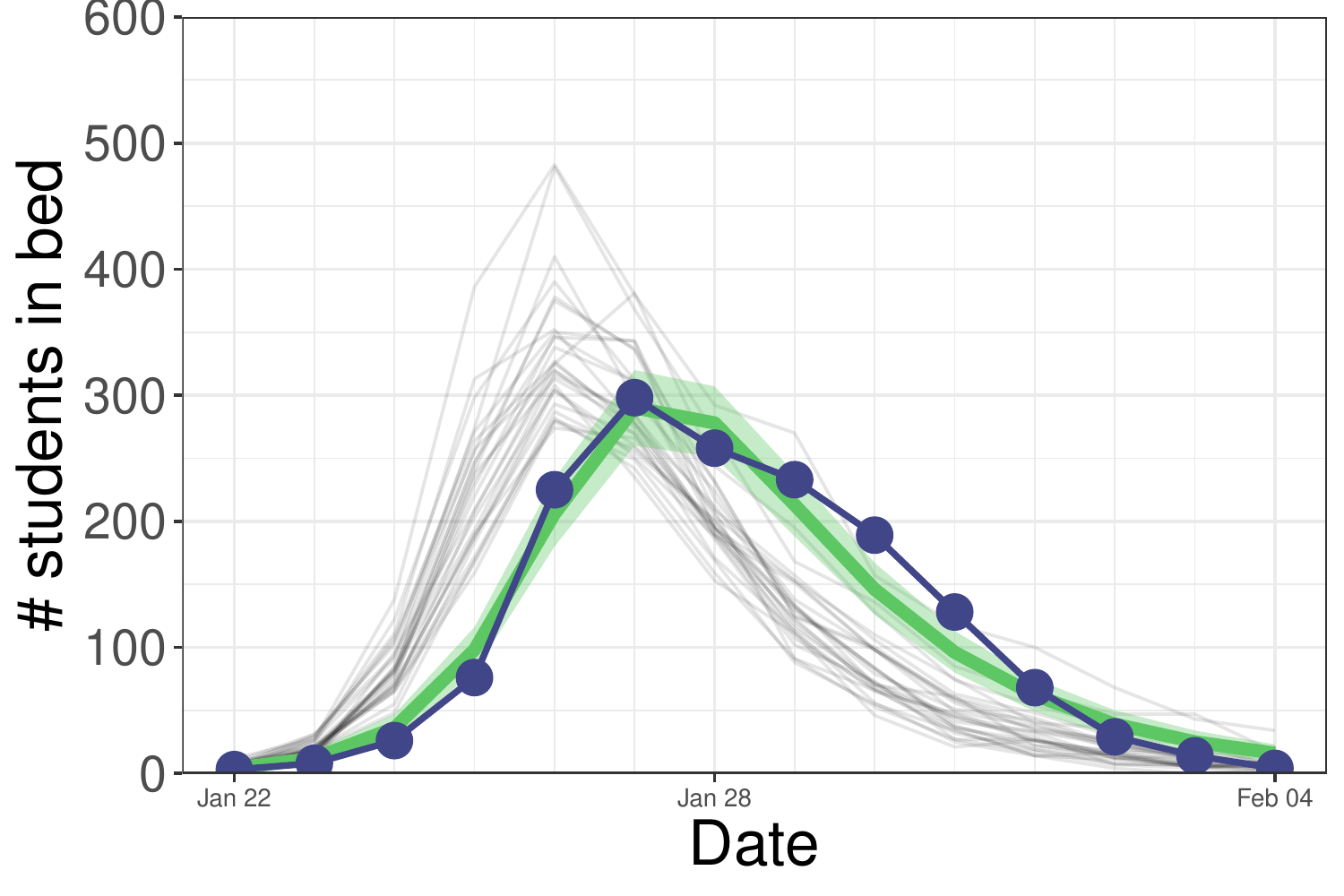}
        \caption{$M_2$: SEIR, Poisson}
    \end{subfigure}
    \begin{subfigure}[t]{0.32\linewidth}
        \includegraphics[width=\linewidth]{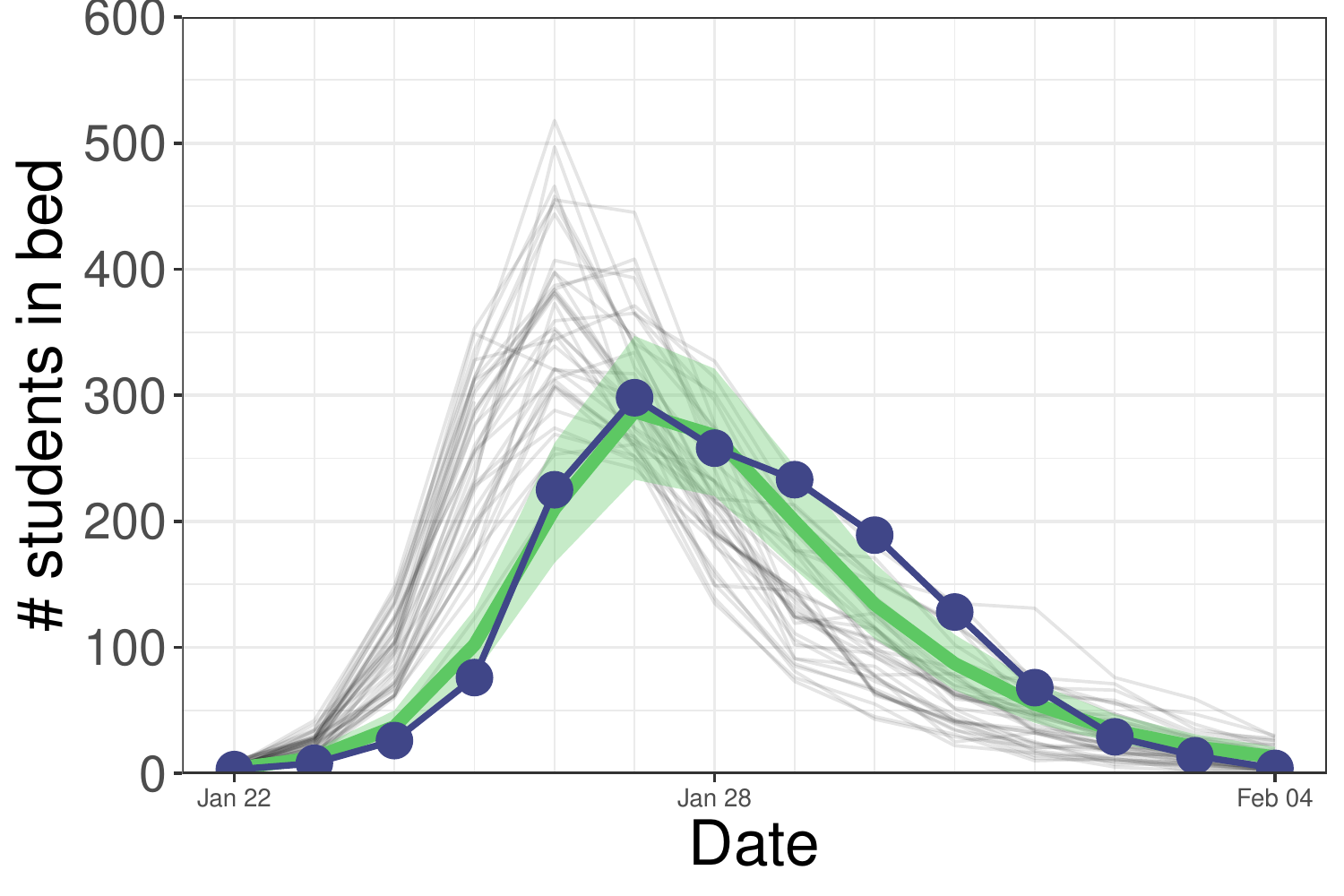}
        \caption{$M_3$: SEIR, Negative-Binomial}
    \end{subfigure}
    \caption{\textbf{Experiment 2.} 
    Simulated data sets~$\{\simulated{y}\}$ from the prior predictive distribution (gray, thin).
    Real observed data set~$\observed{y}$ of interest (blue, thick).
    Posterior predictive distributions~$p(\tilde{y}\given\observed{y}, M_j)$ for each model $M_j$, conditional on the observed data: median (green line) and symmetric 89\% credible interval (green ribbon).}
    \label{fig:experiment-2-posterior-predictive}
\end{figure}

The observed data~$\observed{y}$ from the actual outbreak in 1978 yield the posterior model probabilities~$\observed{\pi}=(0.92, {<}0.0001, 0.08)$.
\autoref{fig:experiment-2-posterior-predictive} shows the posterior predictive distribution of the Bayesian model on the \emph{data}~$y$ (not the meta models, which act on PMPs~$\pi$).
While all posterior predictive distributions heuristically suggest a somewhat good fit, the relatively low number of time steps (i.e., $T=14$) might explain the strong evidence for the simplest model $M_1$ (SIR+Poisson), effectively illustrating Occam's razor.
We hypothesize that longer time series (i.e., more observations) might support more signal, and might thus have the potential to render the more complex models $M_2$ and $M_3$ more likely.

The model-implied PMPs (level 2) show that (i) simulated data from $M_1$ typically shows high PMPs for $M_1$; (ii) data emerging from $M_2$ mostly features low probability for $M_1$, moderate to large PMPs for $M_2$ and moderate probability for $M_3$; and (iii) PMPs for data from $M_3$ either concentrate at extremely large PMPs for $M_3$, or scatter unsystematically across most parts of the probability simplex.
As expected, the meta models (level 3) capture this pattern.

Employing our proposed method in the meta-uncertainty framework, the resulting \emph{predictive mixture} distribution shows high density towards $M_1$.
This is consistent with both the observed PMP ($\observed{\pi}_1=0.92$) as well as the model-implied PMPs under $M_1$, which are usually strongly favoring $M_1$ as well.
The alignment of high observed PMP for $M_1$ on the one hand with the model-implied PMP distribution under $M_1$ on the other hand gives rise to the belief that the observed PMPs are \emph{not} overconfident.
In other words, we expect PMPs for new data sets to feature equally large support for $M_1$, and we believe that the observed PMPs are not a mere artifact of the dance of Bayes Factors \autocite{oelrich_when_2020}.

\clearpage
\subsection{Experiment 3: Simulation-based Inference}

\begin{figure}[t]
    \centering
    \includegraphics[width=0.4\linewidth]{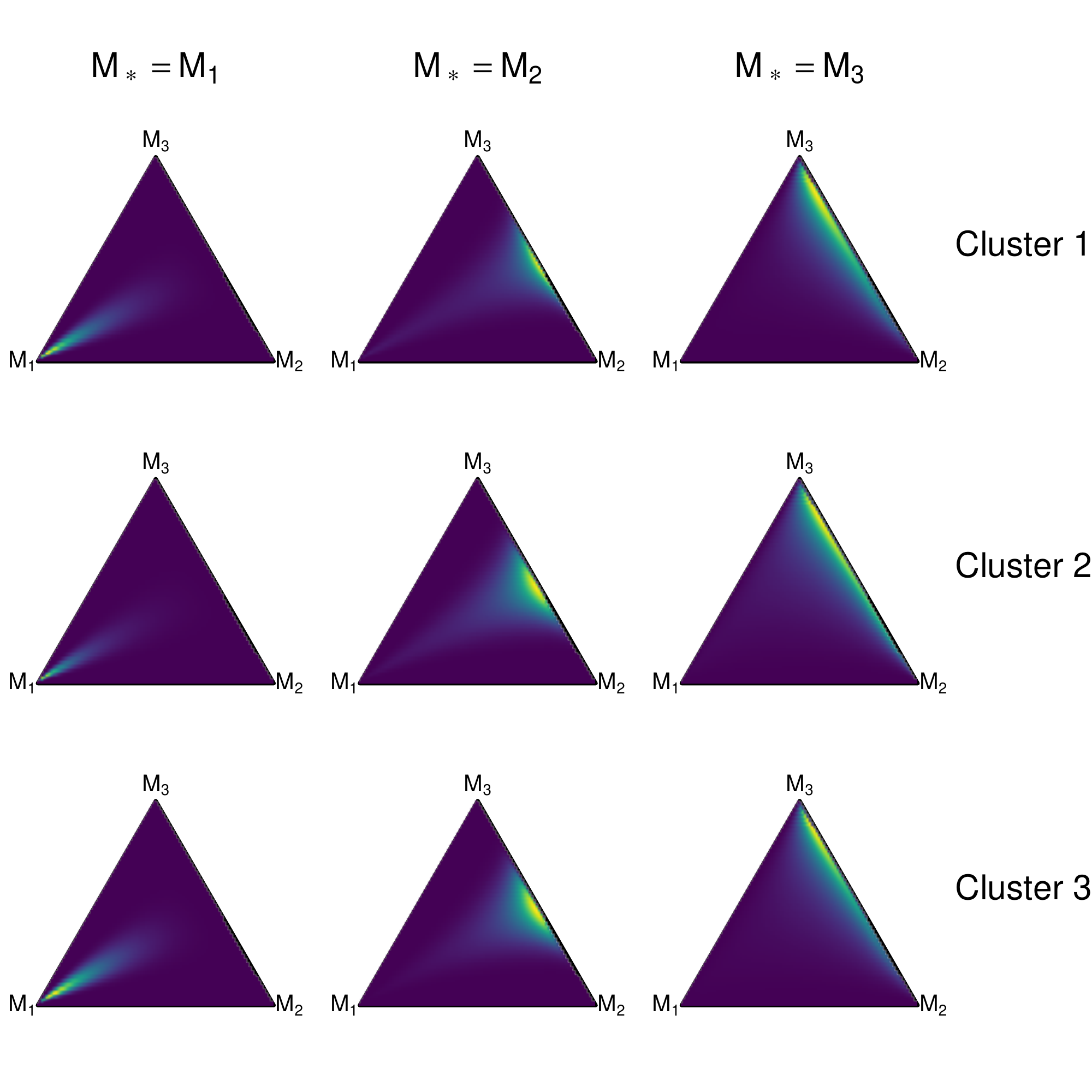}
    \caption{\textbf{Experiment 3.} Pushforward densities of cluster means from the posterior draws of the meta models on simulated data.}    
    \label{fig:experiment-3-clustering}   
\end{figure}

The forward model is adapted from \textcite{radev_amortized_2021} under MIT license. 
It is formulated as a set of five ordinary differential equations (ODEs) describing how the neuron membrane potential $V(t)$ unfolds in time as a function of an injected current $I_{inj}(t)$, and ion channels properties.
The change in membrane potential is defined by the membrane stochastic differential equation (SDE)
\begin{align}
    C \frac{\diff V}{\diff t} &= -I_L - I_{Na} - I_K - I_M + I_{inj} + \sigma \eta (t)
\end{align}
where $C$ is the specific membrane capacitance, $\sigma \eta (t)$ the intrinsic neural noise, and the $I_j$s denote the ionic currents flowing through channels, such that:
\begin{align}
    I_L &= g_L(V - E_L) \\
    I_{Na} &= \Bar{g}_{Na} m^3 h (V - E_{Na}) \\
    I_K &= \Bar{g}_K n^4 (V - E_K) \\
    I_M &= \Bar{g}_M p (V - E_M).
\end{align}
Here, $g_L$ is the leak conductance, while $\Bar{g}_{Na},\Bar{g}_K,\Bar{g}_M$ are the sodium, potassium and
M-type channel maximum conductances, respectively. $E_L,E_{Na}$ and $E_K$ denote the leak equilibrium potential, the sodium
and potassium reversal potentials, respectively. In particular, $g_L$ is assumed constant through time, whilst the other conductances vary over time. Consistently, $(m,h,n,p)$ indicates the vector of the state variables accounting for ion channel gating kinetics evolving according to the following set of ODEs:

\begin{align}
    \frac{\diff i}{\diff t} &= \alpha_i(V)(1-i) - \beta_i(V)i \\
    \frac{\diff p}{\diff t} &= \frac{p_{\infty}(V)-p}{\tau_p(V)} 
\end{align}
where $i\in \{m,h,n\}$, and $\alpha_i(V)$, $\beta_i(V)$, $p_{\infty}(V)$ and $\tau_p(V)$ are nonlinear functions of the membrane potential \autocite[see][ for details]{pospischil_minimal_2008}.

\begin{table}[b]
	\caption{Prior distributions of the parameters in \textbf{Experiment 3}. If a cell contains a scalar value instead of a distribution, the respective parameter is fixed to that value in the model.}
	\label{tab:experiment-sbi-model-priors}
	\centering
	\begin{tabular}{r|cccc}
		& $\overline{g}_{Na}$ & $\overline{g}_{K}$ & $\overline{g}_{M}$ & $g_{l}$\\
		\hline 
		$\M_1$ & $\U(1.5, 15)$ & $\U(1, 3)$ & $0.07$ & $0.1$ \\
		$\M_2$ & $\U(1.5, 15)$ & $\U(1, 3)$ & $\U(0.02, 0.15)$ & $0.1$ \\
		$\M_3$ & $\U(1.5, 15)$ & $\U(1, 3)$ & $\U(0.02, 0.15)$ & $\U(0.05, 0.15)$
	\end{tabular}
\end{table} 

In our simulated experiment, we treat conductances $g_L,\Bar{g}_{Na},\Bar{g}_K$ and $\Bar{g}_M$ as free parameters, and consider different neuronal models based on different parameter configurations. It is also assumed that such configurations allow to affect the span of the possible firing patterns attainable by each model. In particular, we consider 3 models $\mathcal{M} = \{\mathcal{M}_1, \mathcal{M}_2, \mathcal{M}_3\}$ defined by the parameter sets $\theta_1=(\Bar{g}_{Na},\Bar{g}_K)$, $\theta_2 = (\Bar{g}_{Na},\Bar{g}_K,\Bar{g}_M)$, $\theta_3 = (\Bar{g}_{Na},\Bar{g}_K,\Bar{g}_M, g_L)$. 
The parameter priors are listed in \autoref{tab:experiment-sbi-model-priors}.

\begin{figure}[t]
    \centering
        \includegraphics[width=0.3\linewidth]{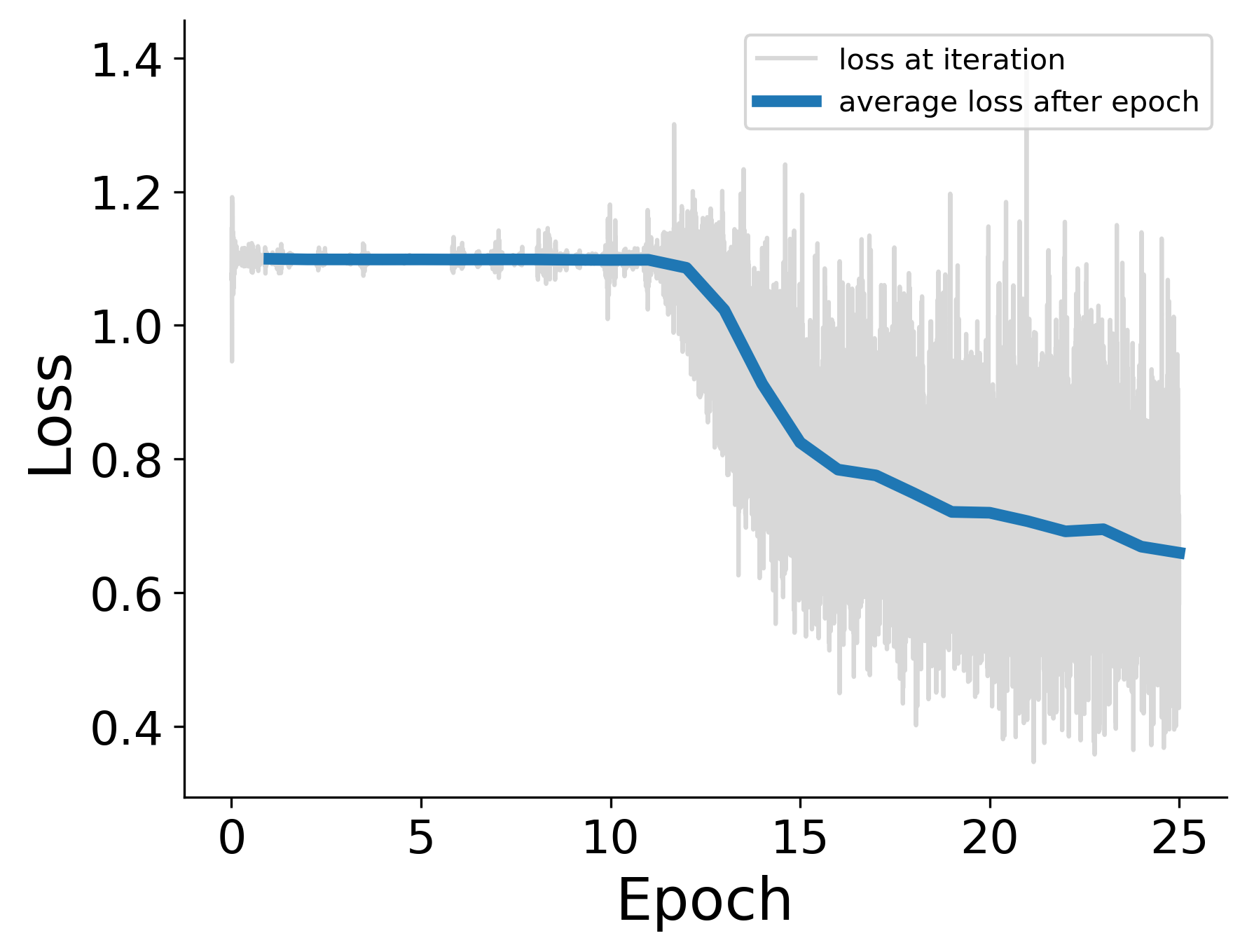}
        \caption{\textbf{Experiment 3.} Loss history of the training phase with BayesFlow.}
        \label{fig:experiment-3-loss}
\end{figure}

For training, we use the Adam optimizer with an initial learning rate of $10^{-4}$. 
We also apply an exponential learning rate decay schedule with a decay parameter $\gamma=0.99$ and a staircase reduction every $1000$ backpropagation steps.
The architecture follows the evidential neural network scheme in \textcite{radev_amortized_2021} with 3 dense layers with ReLU activation and 128 units each, and a softplus activation in the output layer.
The training took 5h on a consumer-grade CPU, with a drastically reduced expected training time on a GPU.
The loss history is depicted in \autoref{fig:experiment-3-loss}.

\clearpage
\FloatBarrier
\section{PROOFS}
\subsection{Proof of Proposition 1.}
\begin{proof}
Assume a closed world setting ($M_*\in\M$) and at least one simulated data set ($K>0$), as well as the regularity conditions \autocite{barron_consistency_1999} that (i) the prior does not put high mass near distributions with very rough densities; and (ii) the prior puts positive mass in Kullback-Leibler neighborhoods of the true distribution.

For an arbitrarily large number of observations $N\to\infty$, the posterior model probabilities $\pi$ (level 1) converge almost surely to one-hot encoded vectors \autocite{barron_consistency_1999}: 
\begin{equation}\label{eq:level-1-consistency}
\pi \xrightarrow{\mathrm{a.s.}} \mathbb{I}_{M_*} \Longleftrightarrow \pi_j = \begin{cases}
    1 & \text{if}\; M_j = M_* \\
    0 & \text{else}
\end{cases}
\end{equation}

For a set of $K$ simulated data sets $\{\simulated{y}^{(k)}\}_{k=1}^K$ with $N\to\infty$ observations each, it follows from \autoref{eq:level-1-consistency} that the resulting set of model-implied PMPs $\{\simulated{\pi}^{(k)}\}_{k=1}^K$ is a set of one-hot encoded vectors $\{\mathbb{I}^{(k)}_{M_j}\}_{k=1}^K$ with a $1$ at the respective true model index and $0$ otherwise.
After grouping $\{\simulated{\pi}^{(k)}\}_{k=1}^K$ by the respective true data generating model, we are left with $J$ groups $G_1, \ldots, G_J$.
Recall that all simulated data sets underlying the PMPs in group $G_j$ stem from the corresponding true model $M_j$.
Thus, according to \autoref{eq:level-1-consistency}, all PMPs in group $G_j$ equal $\mathbb{I}_{M_j}$.
Specifically, the variance of PMPs within each group $G_j$ equals zero, and there is no level 2 uncertainty.

Consequently, fitting a meta-model $\mathbb{M}_j$ on PMPs from group $G_j$ yields a Dirac distribution, which in turn implies no level 3 uncertainty.
Finally, the observed data set $\observed{y}$ features $N\to\infty$ observations as well, implying $\observed{\pi}=\mathbb{I}_{M_*}$ (recall that we assume a closed world).
Thus, both the observed PMPs as well as the mixture components (i.e., meta model predictive distributions) equal $\mathbb{I}_{M_*}$, resulting in a one-hot encoded predictive mixture $f(\tilde{\pi})$ in the limit, which is consistent by definition.
\end{proof}

\subsection{Proof of Proposition 2.}
\begin{proof}
Assume that the predictive mixture does not collapse to a one-hot encoded vector: $f(\tilde{\pi})\neq \mathbb{I}_{M_j}$.
In our framework, this translates to the statement that there is remaining uncertainty at level 2 and level 3.
Given (spuriously) extreme observed PMPs for an arbitrary but fixed model $M_q$, $\observed{\pi}\approx\mathbb{I}_{M_q}$, the predictive mixture reduces approximately to the $q^{\text{th}}$ component: $f(\tilde{\pi})\approx \observed{\pi}_q p_q(\tilde{\pi}\given\{\simulated{\pi}\}_j)$.
Since this component has non-vanishing variance by our initial assumption, the predictive mixture $f(\tilde{\pi})$ has non-vanishing variance as well.
Even if the observed PMPs were one-hot encoded---i.e., $\observed{\pi}=\mathbb{I}_{M_q}$---, the predictive mixture consists exclusively of the $q^{\text{th}}$ component, $\smash{f(\tilde{\pi})=1\cdot \underbrace{p_q(\tilde{\pi}\given\{\simulated{\pi}\}_j)}_{\text{level 2 variance}}}$, which still contains variance from level 2 as per the assumption.
\end{proof}

\end{document}